\documentclass[journal,twoside]{IEEEtran}
\usepackage{amsmath,amsfonts}
\usepackage{algorithmic}
\usepackage{algorithm}
\usepackage{array}
\usepackage[caption=false,font=normalsize,labelfont=sf,textfont=sf]{subfig}
\usepackage{textcomp}
\usepackage{url}
\usepackage{verbatim}
\usepackage{graphicx}
\usepackage{cite}
\usepackage{xspace}

\usepackage{xcolor}
\usepackage{amssymb}
\usepackage[inline]{enumitem}
\usepackage{siunitx}
\usepackage{bm}
\usepackage{dblfloatfix}
\usepackage[edges]{forest}
\usepackage[acronym]{glossaries}
\usepackage{tikz}
\usetikzlibrary{shapes, backgrounds}
\usetikzlibrary{calc}
\usepackage[hidelinks]{hyperref}
\usepackage{orcidlink}

\DeclareSIUnit{\nothing}{\relax}

\newacronym{dsm}{tinyDSM}{tiny Developmental Skill Method}
\newacronym{sdes}{SDES}{Skill Design}
\newacronym{sdev}{SDEV}{Skill Development}
\newacronym{sas}{SAS}{Sensor and Actuator Space}
\newacronym{rl}{RL}{Reinforcement Learning}
\newacronym{kg}{KG}{Knowledge Graph}
\newacronym{ml}{ML}{Machine Learning}
\newacronym{im}{IM}{Intrinsic Motivation}
\newacronym{o_ama}{$o_{AMA}$}{$\mathit{ATOMIC \ MOVE \ [ANGULAR]}$}
\newacronym{o_aml}{$o_{AML}$}{$\mathit{ATOMIC \ MOVE \ [LINEAR]}$}
\newacronym{o_ma}{$o_{MA}$}{$\mathit{MOVE \ [ANGULAR]}$}
\newacronym{o_ml}{$o_{ML}$}{$\mathit{MOVE \ [LINEAR]}$}

\newcommand{\colvec}[2][.9]{%
	\scalebox{#1}{%
		\renewcommand{\arraystretch}{.9}%
		$\begin{pmatrix}#2\end{pmatrix}$%
	}
}
\begin{document}

\title{tinyDSM: A Framework for Skill Modeling and Development for Resource-Constrained Millirobots}

\author{
Markus Kobelrausch\,\orcidlink{0009-0001-8688-8703},
Michael Miedler\,\orcidlink{0009-0005-3607-1171},
and Axel Jantsch\,\orcidlink{0000-0003-2251-0004},~\IEEEmembership{Fellow, IEEE}%
\thanks{Corresponding author: Markus Kobelrausch.}%
\thanks{Markus Kobelrausch and Axel Jantsch are with the Institute of Computer Technology, TU Wien, 1040 Vienna, Austria (e-mail: markus.kobelrausch@tuwien.ac.at; axel.jantsch@tuwien.ac.at).
Michael Miedler was with the Institute of Computer Technology, TU Wien, 1040 Vienna, Austria, when this work was conducted.}%
}

\markboth{}%
{Kobelrausch \MakeLowercase{\textit{et al.}}: \MakeLowercase{tiny}DSM: A Framework for Skill Modeling and Development for Resource-Constrained Millirobots}


\maketitle

\begin{abstract}
In this study, we investigate developmental mechanisms that enable small, resource-constrained systems, such as cm-sized millirobots, to autonomously explore, learn, and adapt their capabilities throughout their lifespans. Reinforcement learning algorithms guide the agent's skill acquisition and adaptation through the interplay of our proposed tiny Developmental Skill Method (tinyDSM), which integrates intrinsic motivation and fitness-based assessment. We strive for minimal hard-wired skills while encouraging the self-directed development of new skills without an externally specified task sequence and within a predefined knowledge graph. A key emphasis in our approach is to encode minimal a-priori general knowledge, which serves as a foundational starting point for the system as it further learns system-specific dependencies from the initial knowledge provided.
Thus, by design, our approach aims to cover generic application domains. The methodology
is based on (a) a developmental mechanism with intrinsic motivation, and
(b) a cognitive architecture (knowledge, reasoning, learning),
while (c) utilizing minimal resources.
It uses a hierarchical knowledge graph and kinematic reasoners to model and evaluate simple and advanced motion-related skills.
In our experiments, we use a resource-constrained millirobot with a volume of \qty{36}{\centi\meter\cubed} with a Raspberry Pi Pico 32-bit microcontroller that integrates all described features and capabilities except the camera system in \qty{9}{\kilo\byte}. Starting with learning the most elementary motor skills, the millirobot autonomously progresses from simple linear and angular movements to complex geometric patterns within 15 minutes. To complement the physical experiments, we perform a simulation-based analysis that enables systematic comparisons across learning algorithms and intrinsic motivation parameters.
\end{abstract}


\section{Introduction}
\IEEEPARstart{D}EVELOPMENTAL robotics investigates mechanisms that enable a robot to continuously explore its environment, learn from its experiences and adapt to changes throughout its lifetime \cite{lungarellaDevelopmentalRoboticsSurvey2003}.
\emph{Cognitive} approaches aim to use a synthetic approach that constructs cognitive functions in a developmentally appropriate manner that is inspired by developmental principles and mechanisms observed in children and animals \cite{cangelosiBabiesRobotsContribution2018a}. Physical embodiment enables the structuring of information through interactions with the environment, whereby the hypothesized developmental model can vary in complexity, ranging from body representation, perception, motor skills and social behavior to linguistic interaction. Such cognitive approaches are studied in the robotics literature at different levels of complexity, with the learning of such models aiming, for example, to acquire skills for navigating a given environment \cite{asadaCognitiveDevelopmentalRobotics2009}. 
Robot learning has received a significant boost from \gls{ml}, with a trend towards advanced robots with methods that process an enormous amount of information and consequently require a lot of resources in terms of memory, time and energy \cite{sooriArtificialIntelligenceMachine2023}. As a result, they require significant amounts of prior knowledge and energy to operate effectively.

\emph{Tiny robot learning} deals with the deployment of \gls{ml} on resource-constrained low-cost autonomous robots. The roots lie in the intersection of embedded systems, robotics, and \gls{ml}. The field faces challenges from size, weight, area, and power constraints, along with sensor, actuator, and compute hardware limitations \cite{neumanTinyRobotLearning2022}. These lightweight robots (weighing under \SI{500}{\gram} \cite{caprariAutonomousMicrorobotsApplications2003}) can operate in small spaces and offer promising solutions for a wide range of applications, from emergency search and rescue \cite{duisterhofTinyRobotLearning2021a} to routine monitoring and maintenance of infrastructure and equipment \cite{derivazInvertedVerticalClimbing2018}. They have limited sensors and actuators and have to learn computationally demanding, complex and robust behaviors in different application spaces. The deployment varies across different robot models, system components, tasks and environments. Consequently, a critical trade-off must be made between energy and memory resources for machine learning and the other system components, such as sensors/actuators and complex behavior.

Our work aims to integrate interdisciplinary fields by addressing the challenge of combining \emph{developmental robotics} concepts within the resource constraints of \emph{tiny robot learning}.
It aims to equip millirobots with a cognitive architecture (limited to knowledge, reasoning and learning) that learns competencies in a development-oriented manner and strives for self-directed learning of new skills and knowledge without an externally specified task sequence, within a predefined knowledge graph. To better comprehend our scope and goals, let us consider the case of a mobile millirobot that aims to acquire motor skills. In this work, we refer to motor skills more generally as motion skills, emphasizing the mobility of the millirobot.

To be very general, we aim at a minimal set of general knowledge that serves as a foundational starting point for the robot with:

\newcommand{\listA}{Knowledge about geometric space, and certain physical relations about friction and motion, as manifest in the kinematic reasoner.}
\newcommand{\listB}{Sensors provide information about the environment, and actuators influence the environment. The exact impact of actuators is unknown, although it is assumed that actuator effects are sensed by sensors (kinematic reasoning).}
\newcommand{\listC}{The concept of skill, as defined. An instinctive desire to explore skills and drive them to high fitness levels, as well as a suitable learning mechanism.}
\newcommand{\listD}{A foundational understanding grounded in a knowledge graph that integrates dependencies for sensors, actuators, and skills.}

\begin{figure}[h!]
\centering
\begin{tikzpicture}[x=1cm, y=0.85cm]

    \def\boxwidth{0.45\textwidth}    
    \def\leftpanel{0.5}  

    \foreach \y/\letter/\step/\desc/\boxheight in {
        0/{A}/{Physical and Geometric Laws}/{\listA}/2.2,
        -2.2/{B}/{Sensors and Actuators}/{\listB}/2.4,
        -4.2/{C}/{Skills, Complex Behaviour, Learner}/{\listC}/2.0,
        -6.0/{D}/{Context Awareness and Knowledge Structure}/{\listD}/1.8
    } {

        \draw[draw=black, rounded corners=5pt, fill=gray!5] (0,\y) rectangle ({\boxwidth}, {\y + \boxheight});

        \draw[thick] ({\leftpanel},\y) -- ({\leftpanel}, {\y + \boxheight});

        \node[font=\bfseries\small] at ({\leftpanel / 2}, {\y + \boxheight / 2}) {\letter};

        \node[anchor=west, font=\bfseries\footnotesize] at ({\leftpanel + 0.2}, {\y + \boxheight - 0.35}) {\step};

        \node[anchor=north west, font=\footnotesize, align=left, text width=0.39\textwidth]
            at ({\leftpanel + 0.2}, {\y + \boxheight - 0.5}) {\desc};
    }
\end{tikzpicture}
\caption{Assumptions and a-priori knowledge that the millirobot has built-in.}
\label{fig:minimalkg}
\end{figure}

Let's assume the millirobot can move using two motors and senses its motion via an acceleration sensor. It knows which kinds of sensors and actuators it has (physical model), but it does not know the effects of actuator commands or anything about the environment - it only has assumptions about sensory information (kinematic reasoner). Initially, the robot queries its knowledge graph based on its available sensor and actuator set and infers competencies that it can potentially develop. We argue that knowledge representation and reasoning are essential as they are universally applicable and allow for generic application or reuse in other contexts \cite{olszewskaOntologyAutonomousRobotics2017}.
As the robot interacts with its environment, it explores its sensorimotor effects, reasons about its motions using its kinematic reasoner, and thus develops an understanding of the relationship between knowledge, perception, and action.
Through that exploration, it develops basic motion competencies, which gradually evolve into more complex skills. It achieves this by engaging in complex behavior where it is \emph{intrinsically motivated} to pursue the most fascinating skills. In psychology, \gls{im} is considered the driving force for autonomous entities to acquire and develop skills, and is the key to their development, as it enables them to effectively deal with problems that arise \cite{ryanIntrinsicExtrinsicMotivations2000}. The robot continuously assesses its learning progress by evaluating its skills' effectiveness (fitness reasoner). This fosters an increasing understanding and expertise, which affects motivation and drives the pursuit of specific skills. Thus, the robot exhibits behavior that strives for continuous development and improvement, while it adapts to any changes in its environment. In the case of an expansion of its knowledge graph, it tackles the challenge of a novel development while maintaining a balance with improving existing skills.

In our experiments, the millirobot develops skills through a hierarchical modeling approach. It begins by discovering movement patterns, stored in the knowledge base, such as angular and linear movements. Subsequently, it learns specific distances and angles based on these patterns. The complex behavior derived from \gls{im} is the focus of this work. We examine various patterns in detail and demonstrate their potential for enabling the development of an agent that operates without a predetermined goal. We analyze the effects of its intrinsically motivated behavior on learning and investigate how it responds to different situations. Additionally, we assess the accuracy of its movements by studying a compound movement pattern, namely a square. To complement these physical experiments, we also perform a simulation-based analysis to systematically evaluate different learning algorithms and intrinsic motivation configurations.

\vspace{1mm}
To this end, we propose \emph{\textbf{\gls{dsm}}}, a framework for millirobots which features
\begin{itemize}
    \item a \emph{developmental mechanism} with \emph{intrinsic motivation},
    \item a \emph{cognitive architecture} (knowledge, reasoning, learning),
    \item all while utilizing \emph{minimal resources}.
\end{itemize}
\vspace{1mm}
We consider this to be a worthwhile approach.
Minimizing prior knowledge and assumptions will facilitate very flexible systems. It will allow the use of sensors and actuators with varying levels of accuracy and may adapt to certain aging and wear-out effects, provided observations remain sufficiently informative about the underlying system state. Our long-term goal is to provide the robot with general methods that enable operation across diverse sensors, actuators, and physical environments. Imagine a wheel-equipped or flying robot operating on level plains, rocky or grassy surfaces, or even in wet environments, and learning any competence that is physically feasible (e.g., if the robot has only LEDs but no motors, it cannot learn to move).

\section{Background}
Our \gls{dsm} is based on several research domains, which we briefly review in the following sections.
The broader domain is developmental robotics, which studies mechanisms that allow robots to learn and adapt independently over time, inspired by developmental principles observed in children and animals \cite{lungarellaDevelopmentalRoboticsSurvey2003,asadaCognitiveDevelopmentalRobotics2009,cangelosiBabiesRobotsContribution2018a}.
Cognitive architectures such as LRMB\cite{wangLayeredReferenceModel2006}, ATC-R \cite{ritterACTRCognitiveArchitecture2019}, or SORA \cite{lairdCognitiveRoboticsUsing} model layered cognition, enabling robots to process perceptual data, reason logically, and organize information into conceptual categories
\cite{levesqueChapter23Cognitive2008}. Our approach is similar to SORA as we work within a decision cycle that includes perception, action, memory and reasoning. We differ from SORA in that we do not rely on comprehensive symbolic reasoning that involves working or long-term memory, but instead use a lightweight knowledge structure and explicit motion-related reasoning. This limitation is intentional, as we want to maintain a minimal cognitive architecture that relies on a minimal set of assumptions.

Robots benefit from a structured semantic model to understand the real world. In particular, ontologies organize knowledge in a way that allows cognitive robots to reason and create understanding \cite{olszewskaOntologyAutonomousRobotics2017}. Knowledge bases are popular in service robots, modeling complex domain-specific knowledge \cite{sunReviewDomainKnowledge2019}. Frameworks such as KnowRob\cite{tenorthKNOWROBKnowledgeProcessing2009}, RoboBrain\cite{saxenaRoboBrainLargeScaleKnowledge2015}, or BWIBots\cite{khandelwalBWIBotsPlatformBridging2017} show success in mastering a variety of complex tasks, ranging all the way to cognitive language skills. KnowRob and RoboBrain focus on knowledge representation with extensive knowledge databases and reasoners for various service-oriented robots. In contrast, our work focuses on a reduced knowledge graph specifically for mobile millirobots. While BWIBots also specializes in navigation, it focuses on a collaborative robot and, unlike us, assumes static task planning. Our approach differs in that we extend to dynamic planning with intrinsic motivation and consider tightly limited resources.

Intrinsic motivation (\gls{im}) is a central concept in lifelong learning and is understood as the drive to engage in activities for their own sake, primarily for enjoyment and satisfaction, rather than because of external rewards or explicit tasks. In robotics, \gls{im} draws inspiration from cognitive science and psychology, where curiosity and self-directed exploration are emphasised in learning \cite{oudeyerIntrinsicMotivationSystems2007,oudeyerWhatIntrinsicMotivation2007,oudeyer:inria-00420175,ryanIntrinsicExtrinsicMotivations2000}. \gls{im} is often combined with Hierarchical Reinforcement Learning (HRL) to create a multi-level policy structure  \cite{bartoIntrinsicallyMotivatedLearning2004,skellyHierarchicalReinforcementLearning,pateriaHierarchicalReinforcementLearning2022,andreasModularMultitaskReinforcement2017}. Our approach is also based on hierarchical competence models but differs in the use of an explicit, structured knowledge graph that the agent actively queries to drive its development. Specifically, we derive a difficulty factor from the graph that is incorporated into the intrinsic motivation. We consider novelty, progress, and difficulty in intrinsic motivation, and our specialized computation demonstrates robust and effective behaviors.

Moreover, we aim to embed learning mechanisms in the highly constrained resource space of tiny robots~\cite{neumanTinyRobotLearning2022}. TinyML offers new possibilities with techniques such as quantization, pruning, and clustering to reduce the computational load and memory requirements of machine learning models, enabling them to be deployed on low-resource devices \cite{immonenTinyMachineLearning2022,leEfficientNeuralNetworks2023,linTinyMachineLearning2023,sahaMachineLearningMicrocontrollerClass2022}. Although this approach is promising, most TinyML systems rely on offline training and therefore do not fully meet our requirements. In terms of algorithms, \gls{rl} is well-suited due to the design of our reinforcement flow in \gls{dsm}. Algorithms such as Deep Q-Networks (DQN) \cite{mnihPlayingAtariDeep2013} and Deep Deterministic Policy Gradient (DDPG) \cite{lillicrapContinuousControlDeep2019} are viable candidates. However, these models are large and require significant resources. Among traditional \gls{rl} algorithms, Q-learning-based algorithms \cite{watkinsQlearning1992a} are effective for low-dimensional search spaces due to their simplicity and low resource requirements.
TinyRL is still in the early stages, but research on deep \gls{rl} in resource-constrained environments is growing \cite{szydloTinyRLReinforcementLearning2022,svobodaResourceEfficientDeep2020a}. Such approaches are promising and offer potential for integration into our framework. Besides that, genetic algorithms also show potential for policy optimization \cite{stanleyEvolvingNeuralNetworks2002,suchDeepNeuroevolutionGenetic2018}.

 Specific work similar to ours includes the research from Baranes and  Oudeyer \cite{baranesActiveLearningInverse2013} who employ an intrinsically motivated approach to learning inverse models by actively selecting goals based on learning progress. Forestier et al.~\cite{forestierIntrinsicallyMotivatedGoal2022} extended this with curriculum-based learning, which allows the complexity of the goals to be self-organized. In contrast, we extend this approach by also considering novelty and difficulty factors in the intrinsic motivation. Moreover, the development frameworks of Nguyen \& Oudeyers \cite{nguyenRobotsLearnIncreasingly2021a} and Colas et al. \cite{colasCURIOUSIntrinsicallyMotivated2019} also focus on the idea of skill-based intrinsic motivation for development.
In summary, our work integrates concepts from developmental robotics, cognitive architectures, intrinsic motivation, and resource-constrained learning to enable on-device skill learning for millirobots.
Our work demonstrates that cognitive architectures can effectively scale to function in highly constrained environments, as confirmed by the minimal resource footprints in our experimental results. Due to the strong growth and interest in IoT devices, we see great potential for our approach to reduce customization efforts in this domain.

\section{Developmental Skill Method}
\subsection{Definition of SAS, Fitness and Skills}
\subsubsection{SAS}
\newcommand{\SAS}{\ensuremath{\text{\bf SAS}}\xspace}
\newcommand{\SASet}[1]{{\ensuremath{\bf\mathcal{#1}}}\xspace}
\newcommand{\Reals}{\mathbb{R}\xspace}

The \emph{Sensor-Actuator Space} $\SAS = (\SASet{S}, \SASet{C})$ is a
2-tuple and consists of a set of sensors
\SASet{S} and a set of actuators \SASet{C}. A sensor reading $s_{i}\in
\SASet{S} \times T \rightarrow \Reals^{P(s_i)}$ is a mapping of a sensor
identifier at a given time to a vector of values read from the sensor
interface. An actuator command $c_j\in \SASet{C} \times T \times
\Reals^{P(c_j)}$ is defined by the actuator identifier $c_j$ and 
$P(c_j)$ parameters that are passed to the actuator interface at a given
time. For instance, for a robot with two motors
$\SASet{C}=\{c_{m_1},c_{m_2}\}$, the two corresponding motor commands
each takes three parameters, the time of activation $t$, the rotational
force $f$ and the duration $\tau$ how long the force is applied:
$c_{m_1}(t,f,\tau)$ and $c_{m_2}(t, f,\tau)$. The robot has one sensor
$s\in \SASet{S}$ that provides the translation information $s(t)
\rightarrow [x,y,\psi]$ at time $t$, with $x$ and $y$ denoting the
translations in the $x$- and $y$-dimensions, respectively, and $\psi$
denoting the yaw angle.

\subsubsection{Fitness}
An integral element of a skill is the assessment of its quality. It is
based on a skill dependent set of quantities. These quantities are
often sensor readings, like distance or velocity, but may include outputs of other skills. For a
fitness that depends on $n$ quantities with $n \in \mathbb{N}, n > 0,$
we define the following vectors.

\begin{equation*}
	\vec{t} = \begin{pmatrix}
		t_{0} \\
		t_{1} \\
		\ldots \\
		t_{n-1}
	\end{pmatrix}
	\\\:
	\vec{o} = \begin{pmatrix}
		o_{0} \\
		o_{1} \\
		\ldots \\
		o_{n-1}
	\end{pmatrix}
	\\\:
	\vec{r} = \begin{pmatrix}
		r_{0} \\
		r_{1} \\
		\ldots \\
		r_{n-1}
	\end{pmatrix}
	\\\:
	\vec{e} = \begin{pmatrix}
		e_{0} \\
		e_{1} \\
		\ldots \\
		e_{n-1}
	\end{pmatrix}
\end{equation*}

\noindent The \emph{target vector} $\vec{t}$ signifies the desired quantities
we seek to reach. The \emph{range vector}  $\vec{r}$ defines the maximum
permissible deviation from the target vector. The \emph{observation vector}
$\vec{o}$ captures the values observed after the execution of an
activity, and the \emph{error vector} $\vec{e}$ is calculated as follows:

\begin{equation*}
	e_i = \frac{|o_i - t_i|}{r_i}, \quad \text{for } i = 0, 1, \dots, n-1
\end{equation*}

The scalar \emph{fitness value} $f \in [0, 1]$ is then computed as:

\begin{equation*}
	f = 1 - \frac{||\vec{e}||}{\sqrt{n}}
\end{equation*}

Putting this together, the fitness $F$ of a skill is a 5-tuple.

\begin{equation}
\label{eq:Fitness}
F=(\vec{t},\vec{r},\vec{o},f(.), f_{\text{threshold}})
\end{equation}

\noindent A skill is deemed ``learned'' if $f(.) \ge f_{\text{threshold}}$. The
required threshold can vary but is typically around $0.95$.
Hence, the fitness function, often called the fitness of a skill, evaluates to a single scalar number
that represents the quality of an activity. It is used by learning
routines to direct the learning process and establish awareness of
the level of competence an agent has at any given time.

\begin{figure}[htbp]
	\centerline{\includegraphics[width=0.23\textwidth]{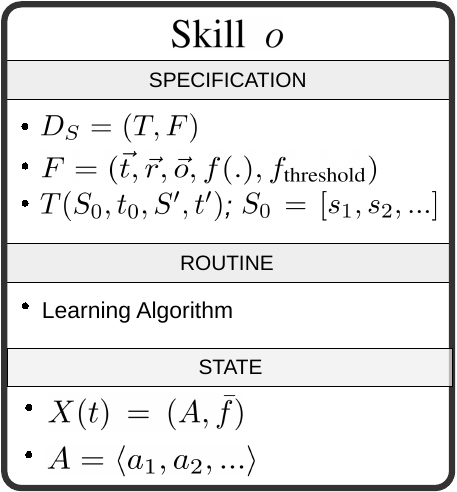}}
	\caption{Overview of the definition of a skill, including its specification, routine and state (during execution).}
	\label{fig14}
\end{figure}

\subsubsection{Skill}
As shown in Fig.~\ref{fig14}, a \emph{skill} $o$ consists of three components: a specification, a learning routine and a state.
The \emph{skill specification} defines the skill in terms of a
relation between the action space and the sensor space. Formally, it is a 2-tuple $D_S=(T,F)$ consisting
of a desired transformation of the sensor readings
$T$ and the fitness $F$, as defined above.
$T(S_0, t_0,
S', t')$ denotes the desired transformation on the sensor readings,
where $S_0=[s_1,s_2,...]$ is the set of sensor readings at time
$t_0$, before the first action $a_1$ is applied, and $S'=[s'_1, s'_2,
...]$ is the set of sensor readings at time $t'$ after the last
action of $A$ has been applied.

The learning routine is a learning algorithm that
modifies the action sequence with the objective of maximizing 
fitness. A reinforcement learning algorithm is a typical and good
example of a learning routine.

The State of a Skill $o$, $X(t)=(A, \bar{f})$ is a 2-tuple
consisting of an action sequence $A$, the current fitness value
$\bar{f}$, as evaluated by the fitness function $f(.)$.
$A=\langle a_1,a_2, ... \rangle$ is
the set of commands $a\in \SASet{C} \times \Reals^{P(a)}$ applied in sequence.
Whenever the skill should be applied, $A$ is executed. Initially, $A$ may be empty
and the learning routine has the task to modify $A$  such that the
fitness is maximized.

We focus on modeling actions that lead to physical actions of the system, but not necessarily motion. For example, consider a robot that has an LED actuator paired with a nearby brightness sensor. When the LED is turned on, it may cause a noticeable change in brightness in the surrounding environment, which the sensor then detects. The objective of the learning algorithm may be to regulate the brightness to a specific target value, enabling the agent to adjust this brightness level based on the correlations it has learned. This scenario exemplifies a lighting skill, with a helper task defining the sequence of brightness actions that signal the SOS distress signal, enabling the millirobot to ask for help.

Note that this concept is general and can extend to all kinds of physical actions.

\subsection{Knowledge Graph}
The foundation of our cognitive system model is a \gls{kg} with semantic features \cite{chenEntitySetExpansion2018}, that contextualizes the system components with dependencies. It reflects the scope of development at a certain point in time. Due to the ability to flexibly extend the \gls{kg}, the robot needs to deal with skills and problems when they arise, which is a challenge to the complex behavior, especially to \gls{im}. Due to limited resources, we seek a simple knowledge graph representation. Currently, the robot can only draw conclusions about its hierarchical relationships with skills and the \gls{sas}. However, recent work shows good progress in building rich socio-physical models for service robots, such as SOMA \cite{danielbesslerFoundationsSocioPhysicalModel2021}.
Knowledge graphs and their hierarchy can accelerate learning, as studied in Curriculum Learning (CL), which is an \gls{ml} technique that conducts training in a meaningful order, from easier to more complex tasks \cite{narvekarCurriculumLearningReinforcement2020}\cite{svetlikAutomaticCurriculumGraph2017}.

\begin{figure}[htbp]
	\centerline{\includegraphics[width=0.5\textwidth]{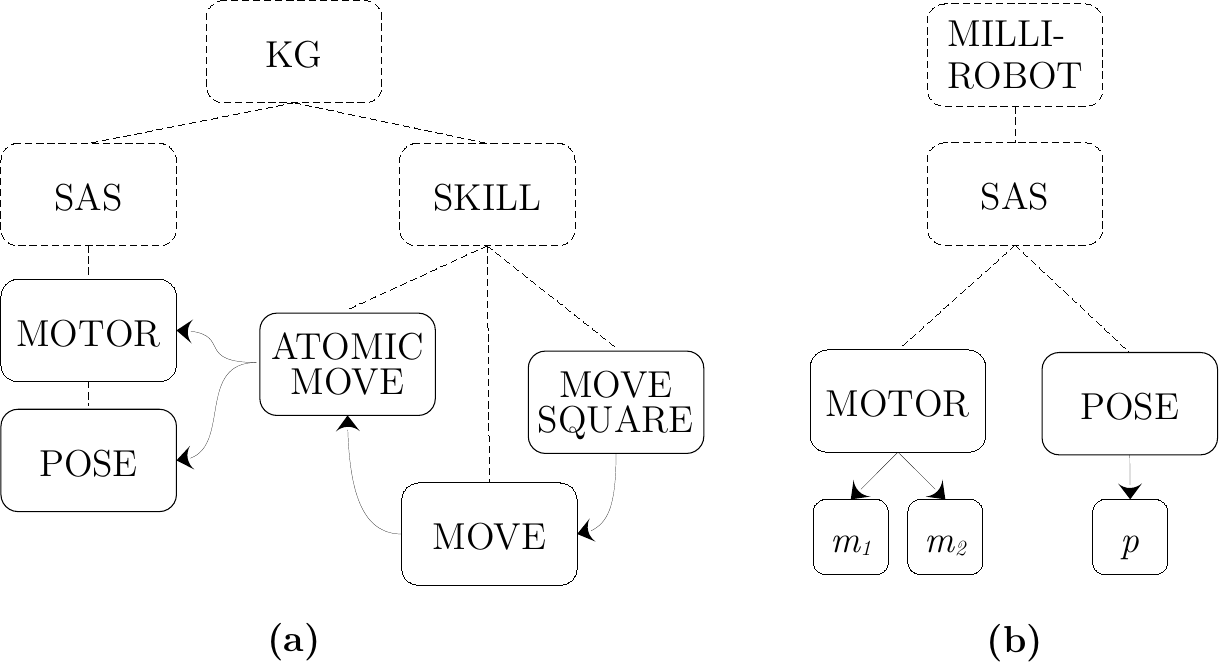}}
	\caption{(a) Generic \gls{kg} of skills and \gls{sas}; (b) Millirobot embodiment defines available sensorimotor capabilities. Skills become masterable once their dependencies are satisfied (e.g., \emph{ATOMIC MOVE} before \emph{MOVE}).}
	\label{fig07}
\end{figure}

In \gls{dsm}, the millirobot uses the generic \gls{kg} to (a) infer the fulfillment of the sensorimotor abilities required for developing particular skills and (b) infer higher-order skills that may become available for development as it evolves.
To represent semantic knowledge, we encode information through types of entities related to skill and the \gls{sas}, as illustrated in Fig.~\ref{fig07}. Let $O$ be a set of skills with $o_i \in O$, \SASet{S} is a set of sensors with $s_i \in \SASet{S}$ and \SASet{C} is a set of actuators with $c_j \in \SASet{C}$, as defined by the \gls{sas}. The set of entity labels (classes/types) follows with $P = O \, \cup \, \SASet{S} \, \cup \, \SASet{C}$. Formally, the \gls{kg}  is a directed acyclic graph $ K = \{E, U, P, \tau\} $, where $E$ is a set of entities, $ U \subseteq \{ (n,m) \, \vert \, (n,m) \in E \times E \land n \neq m \}$ is the set of directed edges and  $\tau: E \to P$ is a bijective function mapping labels to entities. A directed edge $e_{n} \rightarrow e_{m}$ in $K$ between the \gls{sas} and skill entities indicates that a sensor/actuator associated with $e_{m} \in \SASet{S} \triangle \SASet{C}$ needs to be available to the robot's physical body before it can start executing the skill associated with $e_{n} \in O$.
Further, a directed edge $e_{n} \rightarrow e_{m}$ in $K$ indicates that $e_m$ should be learned before $e_n$.
Again, a skill is deemed "learned" if $f(o_{i}) \ge f_{\text{threshold}}$. The set of masterable skills $M_{O}$ at time $t$ includes (a) skills that have been learned and (b) skills that are ready to be learned. Note that, in this work, the \gls{kg} is fixed at runtime, but the framework is designed to support over-the-air updates to its structure and parameters as new capabilities and sensorimotor knowledge become available.

\subsection{Intrinsic Motivation}\label{sec:im}

We model the complex behavior of our millirobot based on its intrinsic motivation. The design aims to generate its own goals and pursue them autonomously, not because they are externally specified in a strict task sequence, but because they appear interesting for various reasons, driven by inner drives and “curiosity” \cite{baranesActiveLearningInverse2013,oudeyerIntrinsicMotivationSystems2007}. According to the taxonomy of \gls{im} models proposed by Oudeyer and Kaplan \cite{oudeyerHowCanWe2009}, our approach aligns most closely with the competence-based models. The millirobot is driven by its desire to improve its skills through a set of self-generated goals instead of primarily seeking novel or surprising stimuli, as often seen in knowledge-based models.
We model the \gls{im} $m(o,t)$ of an agent to pursue a masterable skill with $o \in M_{O}$ as the product of three factors with
\newcommand{\nov}{\ensuremath{\text{novelty}}}
\newcommand{\prog}{\ensuremath{\text{progress}}}
\newcommand{\dif}{\ensuremath{\text{difficulty}}}
\begin{equation*}
	\mathit{m}(o, t) =  \nov(o, t) \cdot \prog(o, t) \cdot \dif(o, t)\\
\end{equation*}
The complete formulation is detailed in Appendix~\ref{appendix:intrinsic-motivation}.

In the context of our work, the robot must be capable of continuous development without stagnation, ensuring it does not get stuck in specific scenarios.
The \emph{novelty} factor reflects how unfamiliar a skill is to the agent. It motivates the robot to explore and discover all skills available in its \gls{kg}, ensuring that no skills are ignored or excluded from learning permanently.
The \emph{progress} factor, primarily driven by the fitness assessment, encourages consistent improvement (fast improvements boost motivation to continue learning) while enabling the millirobot to respond adaptively to regressions.
The \emph{difficulty} factor refers to the complexity of mastering a skill and ensures a balanced exploration, while more complex skills are prioritized. Simpler ones are still in focus periodically and never neglected completely.
With this approach, the millirobot should ultimately be capable of developing all of its skills independently, without an externally specified task sequence and within a predefined knowledge graph.

Addressing the memory and runtime constraints, we seek a lightweight implementation of the \gls{im}, computed using only minimal historical data.

\subsection{Developmental Process} \label{sec:dp}
A key feature of our proposed \gls{dsm} is to encode a minimal set of general knowledge (see Fig.~\ref{fig:minimalkg}) that serves as a foundational starting point for the system. As development unfolds, it learns system-specific dependencies, like sensorimotor mapping skills, monitors them and adapts to new unseen conditions. It is designed to operate within the resource constraints of a low-energy microcontroller with only a few hundred kilobytes of RAM.

\begin{figure}[htbp]
	\centerline{\includegraphics[width=0.38\textwidth]{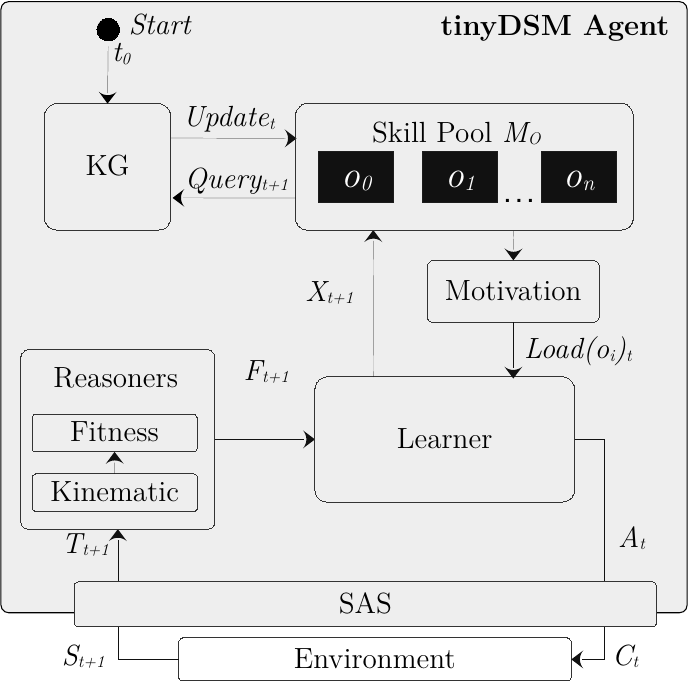}}
	\caption{Overview of the \gls{dsm} developmental flow: the agent queries the \gls{kg} to form a skill pool $M_O$, selects skills via intrinsic motivation, and uses fitness from sensor-based reasoning to drive skill learning and curriculum progression.}
	\label{fig16}
\end{figure}

\gls{dsm} is outlined in Fig.~\ref{fig16} and operates in discrete time steps.
Beginning at time step $t_0$, the millirobot queries its \gls{kg} using its available sensor and actuator set to generate a set of masterable skills, referred to as the \emph{Skill Pool} $M_O$. If the \SAS is not complete, say the millirobot has no motors, it cannot move at all and thus cannot resolve the respective motion-related skills, and its skill pool remains empty.
 However, out of this set, the skill with the highest motivation at time $t$ is loaded for learning. The learner generates a sequence of action commands $A_t$. The \SAS translates them to robot-specific actuator commands $\SASet{C_t}$ leading to physical action in the environment. The robot explores the effects by reading its sensors $\SASet{S_{t+1}}$. The set of reasoners monitors all state transitions $T_{t+1}$ and compares the observed readings to expected assumptions. Based on that, the fitness $F_{t+1}$ is calculated, acting as a reward signal for the learner to refine future actions. Based on $T_{t+1}$ and $F_{t+1}$ the algorithm performs its updates and reports the skill state $X_{t+1}$ to \gls{dsm}. If $f_{\text{threshold}}$ is reached, the skill transitions from active learning to exploitation. In the next discrete time step, the agent re-queries the \gls{kg} and updates the skill pool $M_O$.

At this stage, various scenarios may arise that affect the course of future development: (a) If a skill is deemed learned, it may unlock more complex skills (Curriculum Learning). (b) If the \gls{kg} were extended, it may discover new skills. (c) If the robot undergoes physical modifications, such as adding a new sensor, its developmental space may grow. (d) If none of the above applies, it continues developing existing skills. The curriculum factor in the developmental process directs the agent toward valuable search spaces, maximizing learning efficiency and accelerating progress. In contrast, the intrinsic motivation factor enables autonomous exploration, mirroring adaptive behavior observed in humans and animals. This combination allows the system to rapidly converge to a better local optimum while continuously monitoring and adapting its performance. The agent achieves a structured yet adaptive learning progression by continuously refining its internal representation of learnable skills. This self-reinforcing loop, where knowledge expansion is contingent on prior mastery and environmental modifications, mimics principles observed in biological cognitive development, enhancing both the learning process's autonomy and scalability.

\textbf{Learning routine (skill optimizer)}---
Once tinyDSM has selected a skill for development, a lightweight learning routine optimizes the policy of the skill under strict resource constraints. We are aware of the complexity of optimizing hyperparameters and formulating generalizable rewards, which in most applications require system-specific information to converge quickly. To address this challenge, we use small models with limited completeness and a minimalist reward design in a normalized 0-1 format.
Curriculum learning and knowledge-based reasoners are used to guide the learning process to converge to local optima efficiently.
\gls{dsm} monitors (fitness reasoner) all actions during learning and execution. This enables a transparent learning process and facilitates traceability and validation during execution.
We distinguish between the developmental process implemented by \gls{dsm} (higher-order control logic) and the learning routine (the learner). \gls{dsm} acts as a meta-controller that decides which skill should be learned, when learning should start or stop, and how the skills are organized through the knowledge graph and curriculum. In contrast, the learner is a task-local optimizer that determines how a selected skill is acquired by adjusting its control policy from experience.
Due to resource constraints, we apply simulated annealing (SA) \cite{kirkpatrickOptimizationSimulatedAnnealing1983} for the learner. But any of the well-established \gls{rl} algorithms like Deep Q-Networks (DQN) \cite{mnihPlayingAtariDeep2013} and Deep Deterministic Policy Gradient (DDPG) \cite{lillicrapContinuousControlDeep2019} are viable candidates.
Genetic algorithms also show promise for policy optimization, as evolutionary approaches complement gradient-based methods in overcoming challenges like deceptive local optima and sparse rewards \cite{stanleyEvolvingNeuralNetworks2002,suchDeepNeuroevolutionGenetic2018}.

\section{Experimental Setup}

\begin{figure}[htbp]
	\centerline{\includegraphics[width=0.3\textwidth]{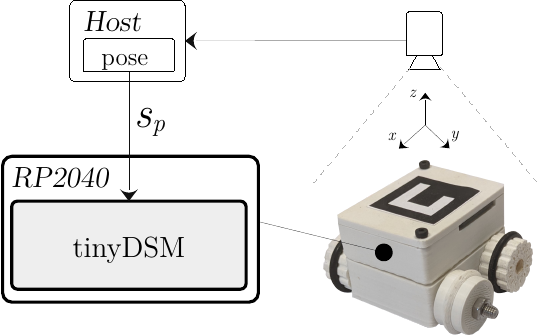}}
	{\caption{The millirobot is 3D-printed and features four wheels, two of which are motor-driven. The translation information is captured on a host device using a camera and a fiducial marker system and transmitted to the embedded RP2040, which operates the \gls{dsm}.}
	\label{fig18}}
\end{figure}

The \gls{dsm} framework is a low-level implementation in C++ that includes several modules, such as memory management, communication (\SAS), skill, scheduler, learner, and agent, among others. Together with an extensive set of low-level libraries that are highly optimized for embedded devices, the framework enables the implementation of various skills and the integration of different learning algorithms, thereby facilitating efficient and flexible studies of developmental mechanisms in low-resource scenarios. In this work, we analyze the memory consumption categorized according to the minimal a priori knowledge and show the runtime performance of the framework without detailing design or implementation aspects. However, the setup for this study is depicted in Fig.~\ref{fig18}, where the millirobot is placed in a controlled environment. A host device (desktop computer) captures the translation information using a camera and a fiducial marker system based on the ArUco library \cite{garrido-juradoAutomaticGenerationDetection2014}.
At this stage, embedded sensors for precise positioning are not yet integrated into the robot, so the host device transmits the sensed pose via Bluetooth. This limits autonomy and restricts experiments to controlled environments. Otherwise, this has no significant impact since there is no major delay in communication, and \gls{dsm} works with any type of position sensor through the generic modeling of the skills, assuming the sensors provide the same types of values. Regarding resources, the communication overhead is about the same as an onboard sensor reading, if not slightly lower. In future work, we will integrate onboard localization to improve autonomy and scalability.
The lightweight millirobot (\SI{150}{\gram}), measuring \SI{40}{}$\times$\SI{30}{\milli\metre}, is 3D-printed and features four wheels, two of which are motor-driven. The \gls{dsm} runs on an RP2040 (Raspberry Pi Pico \cite{raspberrypitradingltd.RaspberryPiPico2021}) \SI{32}{bit} microcontroller, which has tightly limited resources as listed in Table~\ref{tab:rp2040}.

\begin{table}[h!]
\centering
\caption{RP2040 Specifications.}
\label{tab:rp2040}
\renewcommand{\arraystretch}{1.15}
\setlength{\tabcolsep}{6pt}
\begin{tabular}{p{22mm} p{4mm} p{15mm} p{10mm} p{7mm}}
\hline
\textbf{Processor} & \textbf{FPU} & \textbf{HW Integer Divider} & \textbf{Frequency} & \textbf{SRAM}  \\
\hline
ARM Cortex-M0+ & NO & YES & \SI{133}{\mega\hertz} & \SI{264}{\kilo B} \\
\hline
\end{tabular}
\end{table}

For the experiments, we model five motion-related skills. We group them into a foundational $\mathit{ATOMIC \ MOVE}$ skill and a more specialized $\mathit{MOVE}$ skill that builds upon it. These skills enable the millirobot to theoretically follow any geometric pattern, as we demonstrate with the $\mathit{MOVE \, SQUARE}$ skill. As a prospect of a more complex extension, this could also be a navigation path generated by a corresponding $\mathit{NAVIGATION}$ skill.

We investigate how a ``newborn'' millirobot agent discovers these skills based on its physical body and how it develops and adapts them. The skills we propose are hierarchical, meaning that the agent must first master less complex skills before tackling more advanced ones. We examine in detail the intrinsically motivated behavior of the millirobot, which drives its development and must be able to react flexibly and effectively to a variety of unknown situations without stagnating. A key challenge is ensuring the long-term maintenance of skills without neglecting the acquisition of new ones. In addition, we evaluate fitness assessments and demonstrate how they can aid in adaptation to system degradation or environmental changes by adding weight to the robot while operating. In addition to the real-world experiments, we conducted a simulation-based analysis to evaluate learning behavior under varying intrinsic motivation settings (Sec.~\ref{sec:sim-setup}).

\subsection{SAS}
The millirobot's physical model is defined with $\SAS_{robot} = (\SASet{S}, \SASet{C})$. It controls two motors $\SASet{C}=\{c_{m_1},c_{m_2}\}$. The corresponding motor command takes three parameters, the time of activation $t$, the rotational force $f$ and the duration $\tau$, how long the force is applied: $c_{m_1}(t,f,\tau)$ and $c_{m_2}(t, f,\tau)$.
It reads one sensor $s_{p}\in \SASet{S}$ that provides the pose with cartesian coordinate information $s_{p}(t) \rightarrow [x,y,\psi]$ at time $t$, with $x$ and $y$ denoting the horizontal and vertical coordinate in the $x$- and $y$-dimensions, respectively, and $\psi$ denoting the yaw angle.

The fundamental concept is that these interfaces offer an efficient abstraction of the robot-specific sensor readings and actuator commands, enabling higher-level studies of generic methods without being constrained by robot-specific implementations. For example, motor control, whether managed through PWM, needs to be modeled for the specific robotic system. It allows the system to function without needing to know the specific actuators or sensors being used. It can simply rely on the defined interfaces.

\subsection{Kinematic Reasoner}\label{sec:kinematic_reasoner}
For motions, we refer to the basic kinematic and dynamic properties of a system, with kinematics describing the relationship between coordinates in motion space. The dynamics correlate to the torque and force in each wheel. When the wheels touch the ground, these forces act indirectly on the overall system and thus cause it to move. The resulting spatial movement is determined by the change in position over time $t$ using an inverse kinematic reasoner with
\begin{equation}
\label{eq_01}
    \begin{aligned}
        & \Delta x = x_{t} - x_{t+1}, \; \Delta y = y_{t} - y_{t+1}, \\
        & \mathit{\Delta\psi} = \mathit{wrap}(\psi_0, \psi_t) 
    \end{aligned}
\end{equation}
 With $\Delta x$ being the horizontal change in distance, $\Delta y$ is the vertical change, while $\mathit{\Delta\psi}$ denotes the difference in orientation, wrapped to a specified range of \([-180^\circ, 180^\circ]\) with $\mathit{wrap}(\psi_{t}, \psi_{t+1})$. The reasoner utilizes this information to infer (a) \emph{linear} and (b) \emph{angular} motion patterns. A linear movement is defined with the euclidean distance $\mathit{d} = \sqrt{\Delta x^2 + \Delta y^2} \neq 0$ and $\Delta \psi = 0$. While $\mathit{\Delta \psi} \neq 0$ and $d = 0$ indicate a pure angular motion pattern.
  This general knowledge pertains to two-dimensional space and can be used to infer motion for any moving object, in particular, for various robots. Reasoners play a crucial role in knowledge-based systems, enabling logical inference and decision-making from incomplete or structured information. In our design, the kinematic reasoner is particularly characterized, as it adds to the basis of the minimal set of initial knowledge required for the robot to develop. Moreover, an extension to n-dimensional space can be defined in an analogous manner.

\subsection{Skills}

At the foundational level of the knowledge graph, the $\mathit{ATOMIC \ MOVE}$ skill type models basic linear and angular motion patterns. Building upon this, the $\mathit{MOVE}$ skill type models movements over specific ranges, such as traveling specific linear distances or rotating by specific angles. The most complex skill $\mathit{MOVE \ SQUARE}$ uses these lower-level skills to execute a series of movements that outline the shape of a rectangle.

\begin{forest}
for tree={grow=0,  s sep=0pt, edge=thin,  anchor=base west, font=\strut\tiny\sffamily}
[SKILL [ATOMIC MOVE [$o_{AML}$] [$o_{AMA}$] [MOVE [$o_{ML}$] [$o_{MA}$] [MOVE SQUARE [$o_{MS}$]]]]]
\end{forest}
\vspace{2mm}

\noindent\bm{$\mathit{ATOMIC \ MOVE \ [LINEAR]}$}---$o_{AML}$\textbf{:} It is designed to enable the millirobot to drive in a straight line with as little rotation as possible with no specific distance. The specification $D_{S_{AML}}=(T,F)$ includes the desired transformation information $T$ with a set of sensor reading $S=\{s_{p}\}$ where $s_{p} \in T$. As defined by the \gls{sas}, $s_{p}(t) \rightarrow [x,y,\psi]$ provides the cartesian coordinate information at time $t$. The vectors of the fitness $F$ follow with:
\begin{center}
	$\vec{t} =$ \colvec{
        \max (\Delta x , \SI{20}{\milli\metre}) \\
		\SI{0}{\milli\metre} \\
		  \SI{0}{\degree}
    }
	$\vec{o} =$ \colvec{
        \Delta x \\
		\Delta y \\
		\Delta\psi
    }
	$\vec{r} =$ \colvec{
        t_0 \\
	    t_0/4 \\
	    \SI{10}{\degree}
    }
\end{center}
\noindent with the target vector $\vec{t}$, the observations $\vec{o}$ and the permissible deviations $\vec{r}$.
The observation vector $\vec{o}$ captures the sensor readings based on $T$. These are then processed by the kinematic reasoner (see \ref{eq_01}) to determine the relative movement of the millirobot.
During learning, the agent utilizes its algorithm that generates a sequence of motor commands $A = \langle c_{m_1}, c_{m_2} \rangle$ with $c_{m_1}(t) \rightarrow (f_1,\tau_1)$ and $c_{m_2}(t) \rightarrow (f_2,\tau_2)$.
The duration $\tau$, how long the force is applied, is set with a constant value $\tau_1 = C_{\tau}$ and $\tau_2 = C_{\tau}$ to simplify learning, as we search for any straight-line movement regardless of the distance traveled.
Since no specific distance is required, we need to ensure the agent does not learn action commands that are too small. To achieve this, we set a minimum distance of the target $\vec{t}_0$ to \SI{20}{\milli\metre} for the calculation of the fitness score. It still ensures that the motion patterns being explored are independent of the distance traveled. Lateral or rotational movements are not desired. Thus, the target for both is zero. For the same reason as for $\vec{t}$, the members of $\vec{r}$ need to be independent of the distance traveled, thus leaving only forces $f_1$ and $f_2$ to be determined by the learner.

\medskip
\noindent\bm{$\mathit{ATOMIC \ MOVE \ [ANGULAR]}$}---$o_{AMA}$\textbf{:}  It is analogously designed to its linear counterpart $o_{AML}$. It enables the millirobot to rotate in place, without specific angles and with as little movement in the directions $\Delta x$ and $\Delta y$.
The specification is $D_{S_{AMA}}=(T,F)$ with $S=\{s_{p}\}$ where $s_{p} \in T$ and $F$ with:
\begin{center}
	$\vec{t} =$ \colvec{
        \SI{0}{\milli\metre} \\
    	\SI{0}{\milli\metre} \\
    	max(\Delta\psi, \SI{20}{\degree})
    }
	$\vec{o} =$ \colvec{
        \Delta x \\
    	\Delta y \\
    	\Delta\psi
    }
	$\vec{r} =$ \colvec{
        \SI{10}{\milli\metre} \\
    	\SI{10}{\milli\metre} \\
    	t_2
    }
\end{center}
The actuator commands are specified with $A = \langle c_{m_1}, c_{m_2} \rangle$. As no specific angle is given, $\tau$ is also set constant with $\tau_1 = C_t$ and $\tau_2 = C_t$. As above, the learner has to find solutions to $f_1$ and $f_2$.

\medskip
\noindent \bm{$\mathit{MOVE \ [LINEAR]}$}---$o_{ML}$\textbf{:} This skill builds upon $o_{AML}$ to enable the millirobot
to drive in a straight line and stop at a specific distance relative to the starting position. The specification is $D_{S_{ML}}=(T,F)$ with $S=\{s_{p}\}$ where $s_{p} \in T$. Since only the driven distance matters for this skill, the vectors of the fitness $F$ are straightforward:
\begin{center}
	$
	\vec{t} = \begin{pmatrix}
		d_l
	\end{pmatrix}
	$
	$
	\vec{o} = \begin{pmatrix}
		\Delta x
	\end{pmatrix}
	$
	$
	\vec{r} = \begin{pmatrix}
		\frac{d_l}{2}
	\end{pmatrix}
	$
\end{center}
With $d_l$ being the specific distance the robot has to travel. The observations for $\vec{o}$ are calculated by the kinematic reasoner (see \ref{eq_01}) based on the $T$. The learning algorithm generates $A = \langle c_{m_1}, c_{m_2} \rangle$ with $c_{m_1}(t) \rightarrow (f_1,\tau_1)$ and $c_{m_2}(t) \rightarrow (f_2,\tau_2)$. Since the $o_{AML}$ already provides $f_{1}$ and $f_{2}$ for driving in a straight line, the learner solely has to provide the mapping $\tau(d_l)$ where $\tau=\tau_1=\tau_2$.

\medskip
\noindent \bm{$\mathit{MOVE \ [ANGULAR]}$}---$o_{MA}$\textbf{:} Again, it is analogously designed to its linear counterpart $o_{ML}$ and builds upon $o_{AMA}$ to enable rotation by a specific angular distance $d_a$. The specification is $D_{S_{MA}}=(T,F)$ with $S=\{s_{p}\}$ where $s_{p} \in T$.
\begin{center}
	$
	\vec{t} = \begin{pmatrix}
		d_a	\end{pmatrix}
	$
	$
	\vec{o} = \begin{pmatrix}
		\Delta \psi
	\end{pmatrix}
	$
	$
	\vec{r} = \begin{pmatrix}
		\frac{d_a}{2}
	\end{pmatrix}
	$
\end{center}
With $d_a$ being the specific angle the robot has to rotate. The actuator commands are specified with $A = \langle c_{m_1}, c_{m_2} \rangle$.
Since the $o_{AMA}$ already provides $f_{1}$ and $f_{2}$, the \gls{rl} algorithm searches for the mapping $\tau(d_a)$ where $\tau=\tau_1=\tau_2$.

\medskip
\noindent \bm{$\mathit{MOVE \ SQUARE}$}---$o_{MS}$\textbf{:} This skill is a simplified type of skill, as it lacks a learning routine and has $A$ already predefined. Once the skill is invoked, it simply applies $A$. It is used to evaluate all previously learned skills in a compound. The "task" is designed to trace a geometric pattern in a square, which is achieved by a sequence of linear and angular movements. The specification is $D_{S_{MS}}=(T,F)$ with $S=\{s_{p}\}$ where $s_{p} \in T$. The fitness of the geometric pattern is evaluated based on the absolute differences between the global $x$- and $y$-coordinates of the robot's starting and ending points upon completing the task with $F$:
\begin{center}
	$\vec{t} =$ \colvec{
        \SI{0}{\milli\metre} \\
        \SI{0}{\milli\metre} \\
    }
	$\vec{o} =$ \colvec{
        |\Delta x| \\
        |\Delta y| \\
    }
	$\vec{r} =$ \colvec{
        \SI{150}{\milli\metre} \\
        \SI{150}{\milli\metre} \\
    }
\end{center}
The predefined sequence of action commands $A$ follows with:
\begin{equation}
        \left( \; \langle o_{ML}(\SI{50}{\milli\metre}), \; o_{MA}(\SI{90}{\degree}) \rangle \; \right)_{i=1}^{4}
\end{equation}
In this context, $o_{ML}$ signifies a linear displacement of \SI{50}{\milli\metre}, while $o_{MA}$ indicates a rotation of \SI{90}{\degree}, with these movements repeated as necessary to complete the rectangular pattern.

\subsection{Parameterization}
We configure all skills with $f_{\text{threshold}}=0.95$ and the intrinsic motivation as listed in Table~\ref{tab:im-param}.

\begin{table}[h]
\centering
\caption{\gls{im} Parameterization.}
\label{tab:im-param}
\renewcommand{\arraystretch}{1.15}
\setlength{\tabcolsep}{6pt}
\begin{tabular}{p{20mm} p{20mm} p{20mm}}
\hline
$N_{\text{init}}=50.0$ & $N_{\text{limit}}=100.0$ & $\beta=0.1$  \\
\hline
$\gamma=0.003$ & $p_{\text{scale}}=80.0$ & $p_{\text{offset}}=20.0$  \\
\hline
\end{tabular}
\end{table}

\subsection{Simulation-Based Analysis}
\label{sec:sim-setup}

To investigate the developmental dynamics in a controlled and repeatable evaluation, we implemented a physics-based simulation environment in Python using \emph{pygame}. The setup has the same control interfaces and sensor readings (\SAS) as the real millirobot, but differs in certain physical and kinematic properties. The simulation facilitates the development and evaluation of our method, although it does not guarantee direct transfer from the simulation to reality without additional fine-tuning.

We evaluated three learning algorithms:
\begin{itemize}
    \item \textbf{Simulated Annealing (SA)} - used on the real robot,
    \item \textbf{Q-learning} - a Q-Table RL,
    \item \textbf{Random} - a uniform random action selection (as a lower-bound reference).
\end{itemize}
All learners operate under identical intrinsic motivation, fitness thresholds, and knowledge graph structures. The state-action space and reward are the skill definitions and normalized fitness from the section above. For Q-learning we use learning rate $\alpha=0.011$, single-step updates (no discount factor), and $\epsilon$-greedy exploration with $\epsilon$ decaying from 1 to $10^{-3}$. For SA, temperature decays linearly from 1 to $10^{-3}$ with step $3\cdot10^{-4}$. For the intrinsic-motivation configuration analysis, each setting was evaluated over $N=10$ independent simulation runs with different random seeds.

Performance is evaluated with the \emph{mean skill fitness}, defined as the average fitness across the five motion-related skills. This continuous metric provides a compact and sensitive measure of general developmental behavioral abilities.

To characterize the internal dynamics of the intrinsic motivation, we define two metrics:
(i) \emph{selection entropy $H(t)$}, which measures how evenly skills are selected, and
(ii) \emph{maximum neglect $N_{\max}(t)$}, which measures how long any skill remains unselected.
Both metrics are computed from per-skill selection statistics and recency counters (formal definitions are provided in Appendix~\ref{appendix:im-metrics}).

We evaluated six intrinsic motivation parameterizations defined in Table~\ref{tab:im_configs}. These configurations systematically bias the scheduler towards exploration (\emph{high\_explore}), exploitation (\emph{high\_exploit}), higher novelty ranges (\emph{high\_Nlimit}), earlier mastery (\emph{lower\_fthr}), or stronger convergence after the threshold (\emph{high\_postslope}).

\begin{table}[h]
\centering
\caption{IM parameter configurations used in the simulation experiments.}
\label{tab:im_configs}
\renewcommand{\arraystretch}{1.15}
\setlength{\tabcolsep}{4pt}
\begin{tabular}{lcccccc}
\hline
\textbf{IM}
& $N_\text{init}$
& $N_\text{limit}$
& $\beta$
& $\gamma$
& $f_{\text{thr}}$
& $p_\text{post}$ \\
\hline
baseline        & 50  & 100 & 0.10 & 0.003 & 0.95 & 20 \\
high\_explore   & 50  & 100 & 0.10 & 0.010 & 0.95 & 20 \\
high\_exploit   & 50  & 100 & 0.20 & 0.003 & 0.95 & 20 \\
high\_Nlimit    & 50  & 150 & 0.10 & 0.003 & 0.95 & 20 \\
lower\_fthr     & 50  & 100 & 0.10 & 0.003 & 0.90 & 20 \\
high\_postslope & 50  & 100 & 0.10 & 0.003 & 0.95 & 35 \\
\hline
\end{tabular}
\end{table}

To compare IM configurations quantitatively, we define an \emph{IM Score} which combines learning performance with IM scheduling stability. For each run, we compute the time-averaged mean fitness $\bar f$, the time-averaged logarithmic maximum neglect $\overline{N}_{\max}$, and the time-averaged selection entropy $\bar H$. These are aggregated per settings and combined as
\begin{equation}
	\mathrm{IM\;Score}	=
	\bar f -	\lambda\,\overline{N}_{\max}	+ \mu\,\bar H ,
\end{equation}
with $\lambda = 0.25$ that penalizes the long-term skill neglect and $\mu = 0.05$ that rewards the exploration diversity. Fitness and entropy lie in $[0,1]$ and neglect is in log scale. The selected values $(\lambda, \mu)$ result in a ranking that is robust against small fluctuations. For each IM configuration, we show the mean and standard deviation of the IM Score across $n=10$ independent runs in Table~\ref{tab:im_scores} and Pareto analysis in Appendix~\ref{appendix:im-pareto}.

\begin{figure*}[t]
    \captionsetup[subfloat]{font=scriptsize}
    \subfloat[]{\includegraphics[width=1.0\textwidth]{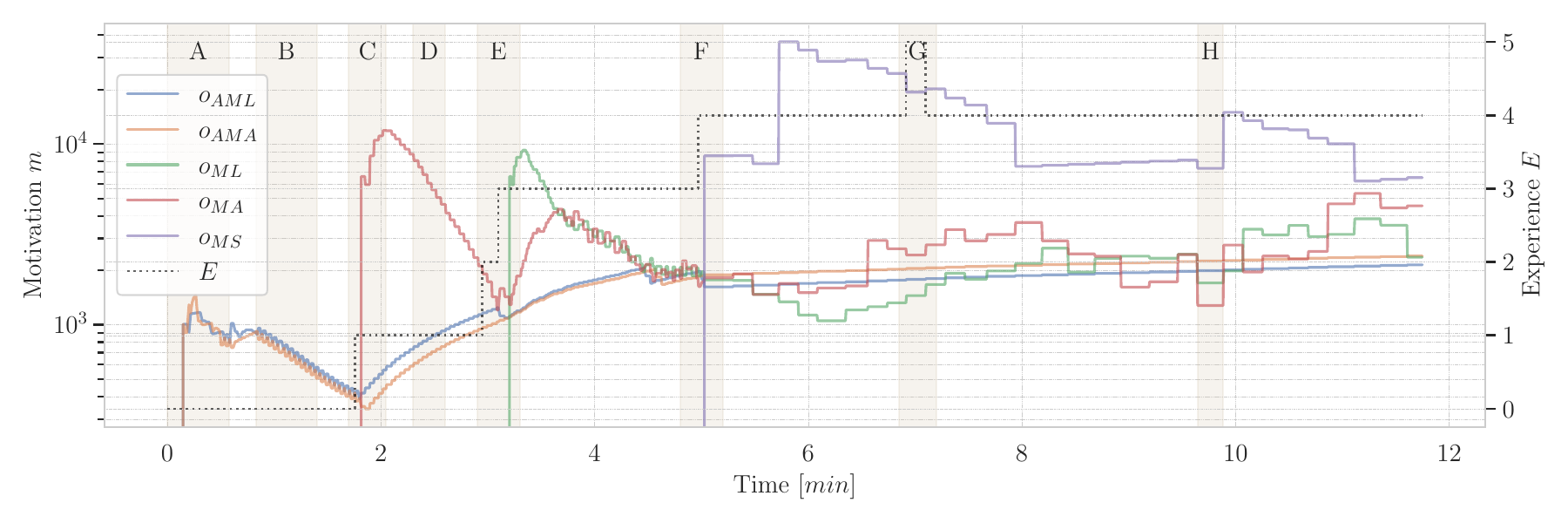} \label{fig:sub_a}} \\
    \subfloat[]{\includegraphics[width=1.0\textwidth]{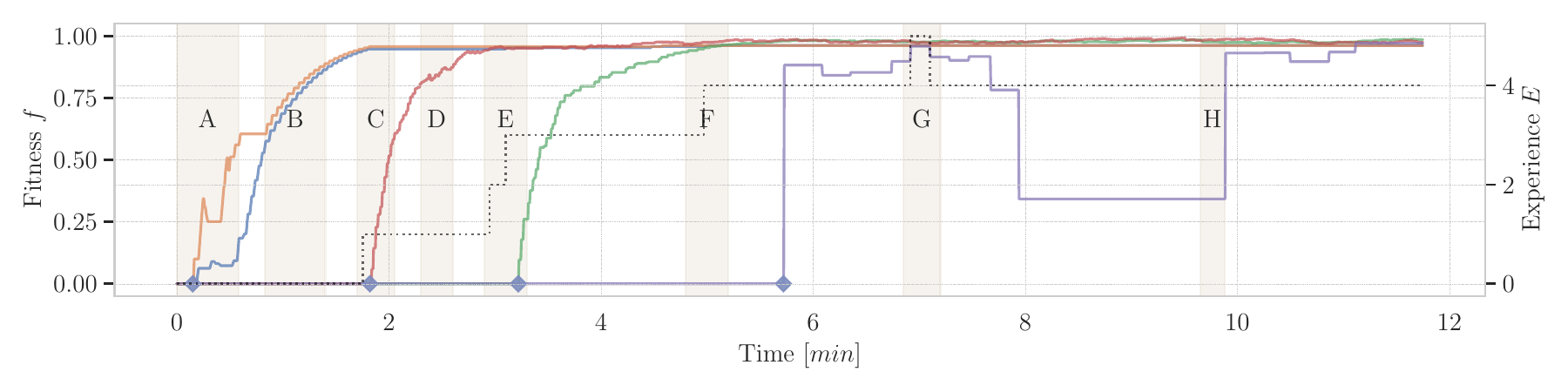} \label{fig:sub_b}}
	\caption{\ref{sec:ex_r_d}: (a) The intrinsic motivation $m$ of the agent for each skill throughout its development, observed over a period of \SI{12}{\min}. The grey segments highlight significant events that we will discuss in the text. (b) The corresponding fitness $f$ for each skill. The blue diamond markers indicate the discovery of new skills. The experience on the right axis of both figures reflects the overall progress of the agent, as defined in Appendix~\ref{appendix:experience}.}
    \label{fig:x_A_01}
\end{figure*}

\section{Experimental Results}
First, in \ref{sec:ex_r_d}, we demonstrate how the millirobot efficiently develops its skills in a reasonable timeframe through its flexible and adaptive behavior. We analyze in detail various factors that contribute to its intrinsic motivation and how these factors influence its actions.
Next, in \ref{sec:ex_r_a}, we explore how the system reacts to environmental changes, such as when the robot is loaded with weight, and how it adapts efficiently in a short time period.
In \ref{sec:ex_r_fa}, we discuss the fitness evaluation in detail, illustrating how it can be applied to various higher-level skills in a scalable manner and the valuable insights we can gain from this process.
In \ref{sec:ex_r_r}, we outline the resources required by our framework, as our \gls{dsm} must operate effectively in target devices with limited resources. Additionally, in \ref{sec:sim-results}, we evaluate the learning behavior and intrinsic motivation dynamics in a simulated environment.

\subsection{Development}\label{sec:ex_r_d}
The millirobot is placed in a controlled environment without developed skills and must learn them from scratch. It starts with its initial prior knowledge as described in Fig.~\ref{fig:minimalkg}.
First, the agent queries its \gls{kg} using its available sensor and actuator set and generates the skill pool $M_O$.
In Fig.~\ref{fig:x_A_01}b, the blue diamond markers indicate when the agent's skill pool is expanded.
Shortly after the start, around \SI{100}{\milli\second}, only $o_{AMA}$ and $o_{AML}$ are available, as they do not have dependencies other than on the \SAS. This query process occurs at each discrete time step, which may lead to an expansion of the skill pool when lower-order skills reach their $f_{\text{threshold}}$ and higher-order skills are available in the \gls{kg}. If a new skill is discovered, the agent may start developing it. This behavior demonstrates the \emph{curriculum learning} element of \gls{dsm}, which guides the agent during initial learning phases to explore valuable search spaces. Once a skill is discovered, the agent's behavior is driven by its \emph{intrinsic motivation}, illustrated in Fig.~\ref{fig:x_A_01}a. It is calculated based on three factors (\ref{sec:im}): \emph{novelty}, \emph{progress}, and \emph{difficulty}. Each of these factors influences the agent's behavior depending on its current state and the experiences gained during the development process.

In Fig.~\ref{fig:x_A_01}, eight relevant segments are highlighted and labeled (A) to (H).
In segment (A), the motivation for both types of atomic motions increases sharply due to the novelty of these skills. They are being discovered for the first time, with no prior experience available. The millirobot randomly selects one skill since both motivations are equal. The agent starts by exploring $o_{AMA}$. The skill $o_{AML}$ also receives sporadic attention, although faster progress is made with the angular pattern, as indicated by increases in fitness (Fig.~\ref{fig:x_A_01}b (A)).

Between segments (A) and (B), the millirobot catches up with the progress of $o_{AML}$. In (B), the development of both skills occurs alternately, as the growth in fitness shows significant similarities. Other experiments have shown that motivation sometimes exhibits a "latching" behavior, focusing on a specific skill for an extended period before switching to the other. In such cases, the fitness of the focused skill increases significantly faster than that of the other skill. However, during (B), the motivation for both skills decreases, even though fitness continues to grow. This decline in motivation is attributed to the novelty and fitness, as the agent becomes more competent and "bored".

In (C), after around \SI{1.8}{\min}, the millirobot successfully reached the $f_{\text{threshold}}$ for $o_{AMA}$, resulting in experience (defined in Appendix~\ref{appendix:experience}) increasing to $1$. This leads to the discovery of $o_{MA}$ and an expansion of the skill pool. Achieving this took only about $\sim50$ interactions with the physical environment, which is fairly efficient. $o_{MA}$ is novel, and the agent is motivated to explore it. That skill is also more complex, as it needs the $o_{AMA}$ for its activities. This is modeled with the higher hierarchical order in the \gls{kg}, which results in a greater difficulty factor that increases motivation further.

In segment (D), the agent focuses only on developing $o_{ML}$. While its fitness grows and it progresses with that skill, the motivation decreases due to the decreasing novelty. This is the same pattern as in segment (B) and will continue to appear throughout development. What is new in this phase is that the agent becomes more interested in rediscovering the two atomic skills, so the motivation for those slightly increases again. However, the agent is slightly more motivated to pursue $o_{AML}$ since it considers itself more experienced with $o_{AMA}$.

In segment (E), the agent faces a new situation as it progresses with $o_{MA}$ and reaches $f_{\text{threshold}}$, driving its experience to $2$. It briefly keeps its focus on angular motions before facing $o_{AML}$ again. This is delayed due to the higher difficulty factor. However, the agent reaches the $o_{AML}$ threshold in the next interaction, increasing its experience to $3$. That leads to the discovery of $o_{ML}$. Shortly after a few interactions, the agent focuses on this newly discovered skill, repeating the pattern observed in (C) and (F).

In segment (F), $o_{MS}$ is discovered, and the millirobot is motivated to drive on a square path, exploiting all previously learned skills in a sequence. However, the agent focuses on navigating along a square for the rest of the experiment. This behavior aligns with the intended objective, as more complex skills typically offer greater utility and adaptability than simpler ones. Even though it focuses on more complex skills, it does not entirely ignore simpler atomic moves, visible in segment (H).

In a brief segment (G), the robot successfully executed some high-quality squares indicated by toggling the experience to $5$. Note that the $o_{MS}$ has no learner, so the agent cannot improve it directly. We discuss this aspect in more detail in experiment \ref{sec:ex_r_fa}.
Finally, the fitness is monitored for each interaction and skill, with the agent immediately recognizing environmental changes. As a result, the agent must be able to adapt to these changes. This challenge will be analyzed in experiment \ref{sec:ex_r_a}.

In summary, in this experiment, we demonstrated that the millirobot can develop skills efficiently and without stagnation in a relatively short time, utilizing \gls{dsm} mechanisms and driving the development by combining \emph{intrinsic motivation} and \emph{curriculum-based} learning. Notably, the millirobot learned basic movements in only $\sim\SI{4.5}{\min}$ and could, in principle, navigate effectively with the respective knowledge acquired.

\subsection{Adaptation}\label{sec:ex_r_a}
\begin{figure}[htp]
	\centerline{\includegraphics[width=.5\textwidth]{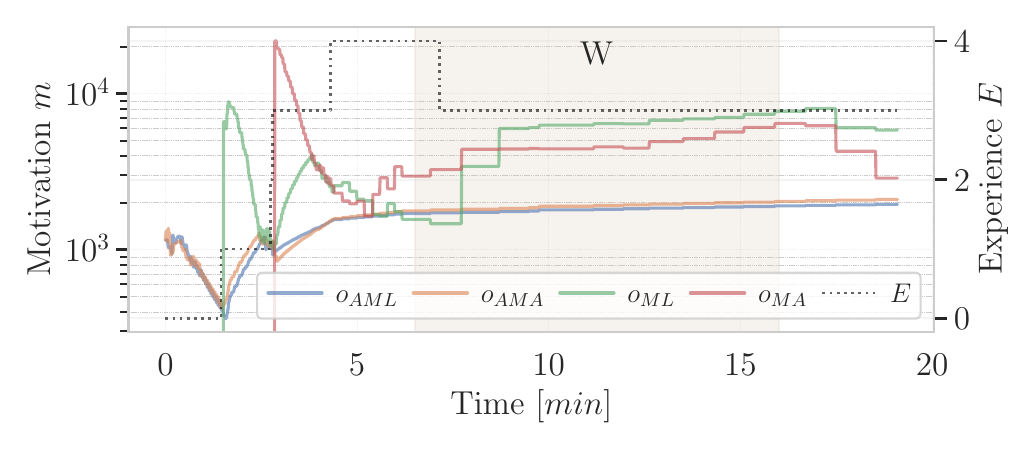}}
    \centerline{\includegraphics[width=.5\textwidth]{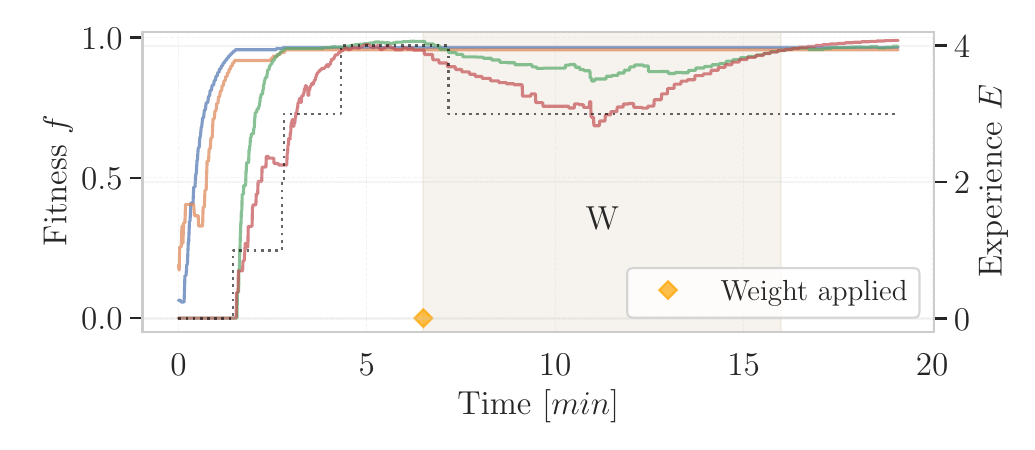}}
	\caption{\ref{sec:ex_r_a}: The upper graph shows the intrinsic motivation $m$ of the agent. The lower graph shows the corresponding fitness $f$ for each skill. The dashed line illustrates the agent's experience. Once the development has progressed and the robot has experience, it is loaded with an additional weight of \SI{380}{\gram}. This is roughly after \SI{6.5}{\min}, which is indicated by the orange diamond marker. This results in a significant drop in fitness values, which motivates the agent to relearn the affected skills during (W).}
    \label{fig:ex_adaption}
\end{figure}

Next, we investigate how the millirobot reacts to environmental changes. We begin by placing the robot in a controlled environment without developed skills.
Fig.~\ref{fig:ex_adaption} shows the progress of the robot. After roughly $\sim\SI{6.5}{\min}$, we load it with additional weight. This is indicated by the start of the gray-shaded segment (W). This alteration in its physical properties invalidates the previously learned movement parameters, as the millirobot now reacts differently to the same motor commands.

Shortly after this weight change, we observe a significant drop in the fitness values of both $o_{MA}$ and $o_{ML}$. Additionally, the agent's experience decreases as its skills drop below the fitness threshold. Note, the weight change only affects the learned duration of the motor commands $\tau$ and does not impact the ratios between $f_1$ and $f_2$. Consequently, the atomic motion skills $o_{AMA}$ and $o_{AML}$ remain unaffected.

However, the decrease in fitness results in a sharp increase in motivation, primarily caused by the progress factor that drives the robot's interest in relearning these skills. During (W), which lasts for $\sim\SI{9}{\min}$, the agent successfully relearns its skills, as evidenced by increasing fitness scores and improvement in experience.

In summary, through the use of \gls{dsm}, the agent continuously monitors all actions, even if the skills are sufficiently developed. It can detect deviations from learned actions through fitness computations. Ultimately, this allows the agent to respond effectively with its behavior to new and unseen situations, adapting its skills to different conditions.

\subsection{Fitness Assessment}\label{sec:ex_r_fa}
\begin{figure}[htbp]
    \centering
    \subfloat{\includegraphics[width=0.5\textwidth]{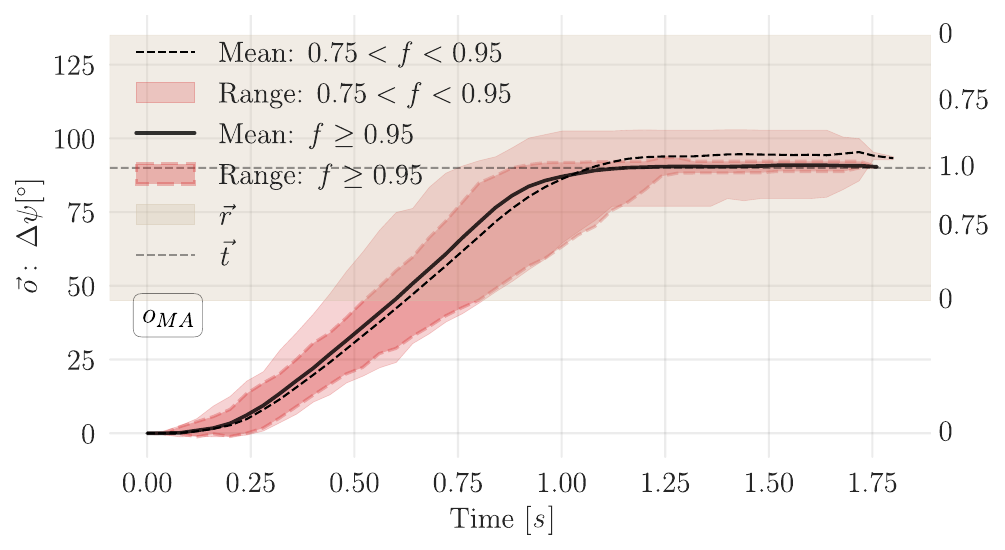} \label{fig:x_r_fa_ma}} \\
    \subfloat{\includegraphics[width=0.5\textwidth]{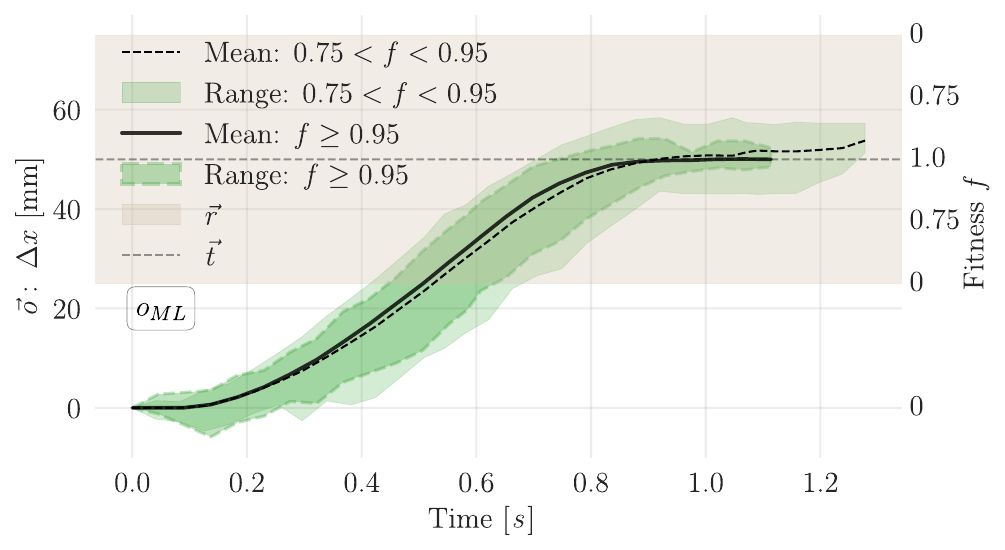} \label{fig:x_r_fa_ml}}
	\caption{\ref{sec:ex_r_fa}: Fitness evolution of angular ($o_{MA}$, red) and linear ($o_{ML}$, green) motion skills during learning. 	Observations $\vec{o}$ (left axis) are compared to targets $\vec{t}$, with fitness $f$ shown on the right. Two learning stages ($0.75<f<0.95$ and $f\geq0.95$) show reduced variance as skills develop.}
    \label{fig:x_f_m}
\end{figure}

Fig.~\ref{fig:x_f_m} illustrates the fitness of $o_{MA}$ and $o_{ML}$ during learning in two distinct progress stages. Movements with less learning progress ($0.75 < f < 0.95$) show a very high variance and deviate significantly from the target, while more developed movements ($f \geq 0.95$) are much closer to the target and show a significant reduction in variance. However, some variance remains even at high fitness values. We attribute this to the physical properties of our millirobot, which is far from ideal.

\begin{figure}[htp]
    \centering
    \begin{minipage}{0.24\textwidth}
        \centering
        \includegraphics[width=\textwidth]{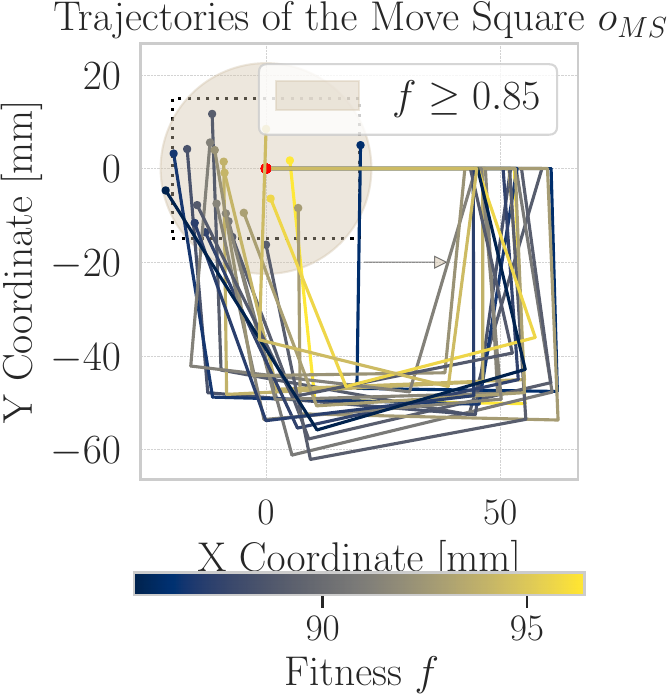}
    \end{minipage}
    \hfill
    \begin{minipage}{0.24\textwidth}
        \centering
        \includegraphics[width=\textwidth]{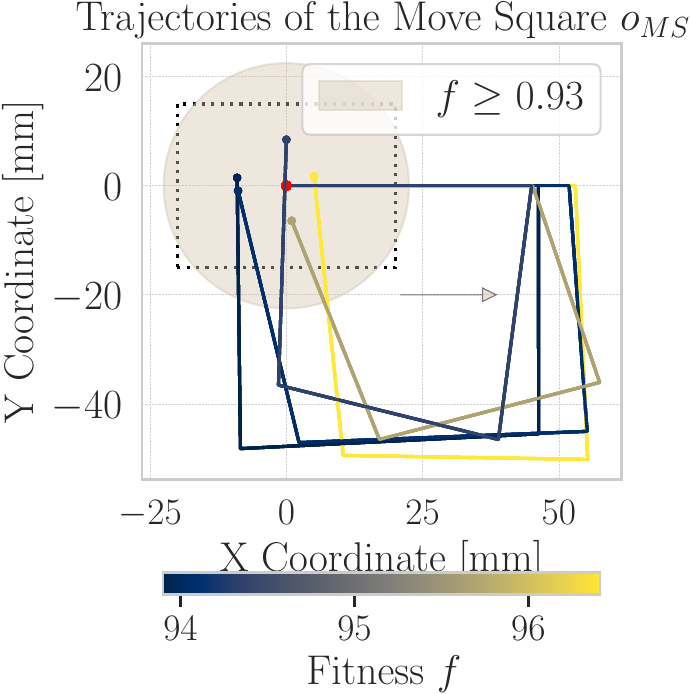}
    \end{minipage}
	\caption{\ref{sec:ex_r_fa}: The Square motion skill $o_{MS}$ projected with the first segment horizontal. The robot starts at the red point and executes alternating \SI{50}{\milli\metre} linear and \SI{90}{\degree} angular moves to form a square. Left: $f\geq85$; right: $f\geq93$.}
	\label{fig:x_f_s}
\end{figure}

During the experiments, we observed that the robot sometimes reacts differently to the same motor commands. Particularly noteworthy is that it experiences a “push” in the braking phase before it stops. This phenomenon is due to the unpredictable stalling of the motor's gearbox when turning off. This results in an extended movement and can cause the robot to overshoot the target. This is also evident in Fig.~\ref{fig:x_f_s}, where, on average, all squares remain clearly “open” and miss the target on the left. Only a few good outliers extend beyond the target to the right. The trajectory deviates from a perfect square shape and does not close exactly at the endpoint, as observed in the significant fluctuations in the fitness calculation of the dependent motion skills.

However, we do not aim for a highly optimized control system but rather want to explore a general approach to learning movement.
The cumulative navigation error over $8$ movements along a square path averages \(\sim\SI{7.5}{\milli\metre}\) over $10$ runs. Since this experiment is actually an open-loop control system, it provides a pessimistic bound on error accumulation. We consider this in the context of our future goal, in which the millirobot with DDQN navigation will be trained to reach a target area with a radius of \(\SI{50}{\milli\metre}\) within a limited working area of \(880 \times 580\,\si{\milli\metre}\). In the intended system, a target-conditioned DDQN selects movements in a reinforcement loop and reevaluates the state after each action, so that motor, friction, and battery-related deviations occur as limited disturbances that the policy learns to compensate for. Even a conservative worst-case error, which is twice as high, remains well within the target range, supporting the statement that the cumulative averaged error of \(\sim\SI{7.5}{\milli\metre}\) is sufficient for reliable target-oriented navigation under the planned constraints.

In summary, we demonstrate that our fitness calculation scales properly and seamlessly applies to more complex skills. The hierarchical structure is also well reflected in the fitness calculation of the square. Errors at lower hierarchical levels directly affect higher levels. By recognizing these errors at different levels, we can correct deviations in a targeted manner. This leads to a significant increase in efficiency, as only the skills that are actually affected need to be relearned, as demonstrated with the adaptation experiment.

\subsection{Resources} \label{sec:ex_r_r}

In terms of resources, we consider the memory and execution time of the whole runtime system during the specific scenarios in our experiments. Since \gls{dsm} was designed for flexible use, we aim to reflect this flexibility in its implementation.

\begin{table}[htbp]
\centering
\caption{Static memory usage per module, including total allocation.}
\label{tab:memory-usage}
\renewcommand{\arraystretch}{1.15}
\setlength{\tabcolsep}{6pt}
\begin{tabular}{p{25mm} p{20mm} p{15mm}}
\hline
\multicolumn{3}{c}{\gls{dsm}} \\
\hline
\textbf{Module} & \textbf{Memory} & \textbf{Usage} \\
\hline\hline
Agent & $\SI{280}{\byte}$ & $5.7\%$ \\
Skills & $\SI{488}{\byte}$ & $10.0\%$  \\
Learners & $\SI{2544}{\byte}$ & $52.1\%$ \\
\SAS & $\SI{840}{\byte}$ & $17.2\%$ \\
Scheduler & $\SI{728}{\byte}$ & $15.0\%$ \\
MM & $\SI{4017}{\byte}$ & $-$ \\
\hline\hline
\textbf{Total} & $\SI{8897}{\byte}$ & $100\%$ \\
\hline
\end{tabular}
\end{table}

\gls{dsm} is designed to only allocate memory during startup and skill creation and never during the normal execution cycle. This makes the memory consumption very predictable, minimizes bookkeeping for the memory manager and reduces the possibility of out-of-memory errors.

Table~\ref{tab:memory-usage} illustrates the static memory consumption. It details the memory usage for each module. The five skills  tested in the experiments require \SI{488}{\byte}, while the communication module (\SAS) uses \SI{840}{\byte} and the scheduler (\gls{im}) requires \SI{728}{\byte}. However, these compact modules represent only a small part, with learners consuming a significant $52\%$ of the total memory. Learning algorithms, particularly neural networks, are memory intensive. Therefore, we focused on a memory-efficient implementation when designing the framework to ensure there is enough space for these algorithms. In this particular case, the numbers for the learners are inflated by a factor of $\approx4$. This is because of internal memory fragmentation caused by the buddy system having a minimum allocation size of \SI{64}{\byte} and therefore being unsuited for many very small allocations. Additionally, the use of dynamic arrays in learners for arrays that are effectively constant in length and contain only one to two entries leads to a significant increase in memory consumption.

As an outlook, we estimate how many skills could be located on the RP2040 in a fictitious scenario. We scale the framework to utilize the full \SI{250}{\kilo\byte}(\SI{16}{\kilo\byte} reserved for pico SDK) available memory of the Pico. We assume one of the move skills for the workload calculation, as this is the most complex and requires the most memory, say the linear move skill. Based on that, our calculations show that $96$ instances of that skill would fit in the pico. This means that the agent would learn $96$ separate linear movements, which is not a useful task but demonstrates \glspl{dsm} potential.

In terms of resource balancing, let's consider a successfully developed skill (threshold reached). In this case, the framework could release the memory allocated for the particular learning algorithm. This leads to a dynamic adjustment of resources, allowing the millirobot to balance its workload according to its needs.
For instance, in segment (H) (see Fig.~\ref{fig:x_A_01}b), no learners are required since all skills have been learned, and $o_{MS}$ has no learner. The memory of \SI{2544}{\byte} (Table~\ref{tab:memory-usage}) can be freed. The robot can still explore its environment and search for new interesting skills, but it requires less energy than in its development phase. This property is ideal for operating in low-resource environments.

\gls{dsm} uses a dynamic Memory Management (MM), which is implemented using a custom buddy memory allocator. Where each module manages its own memory region, sometimes using its own allocation techniques.
One example is the use of a simple memory arena or bump allocator in the skills module, which almost completely removes internal memory fragmentation within this module.
This principle could also be applied to the other modules, reducing memory usage even further.

As for the other resources, the execution time of a single learning step varies from \SIrange{10}{30}{\milli\second} depending on the complexity of the skill. In terms of weight, the millirobot itself weighs \SI{150}{\gram} and the additional weight used in the experiments is \SI{380}{\gram}. During all our experiments, the robot operated for about $\sim\SI{45}{\min}$ on a battery with a capacity of \SI{250}{\milli\ampere\hour}.

\subsection{Simulation-Based Analysis}
\label{sec:sim-results}

\subsubsection{Skill Acquisition Across Learning Algorithms}
\begin{figure}[htbp]
	\centerline{\includegraphics[width=0.5\textwidth]{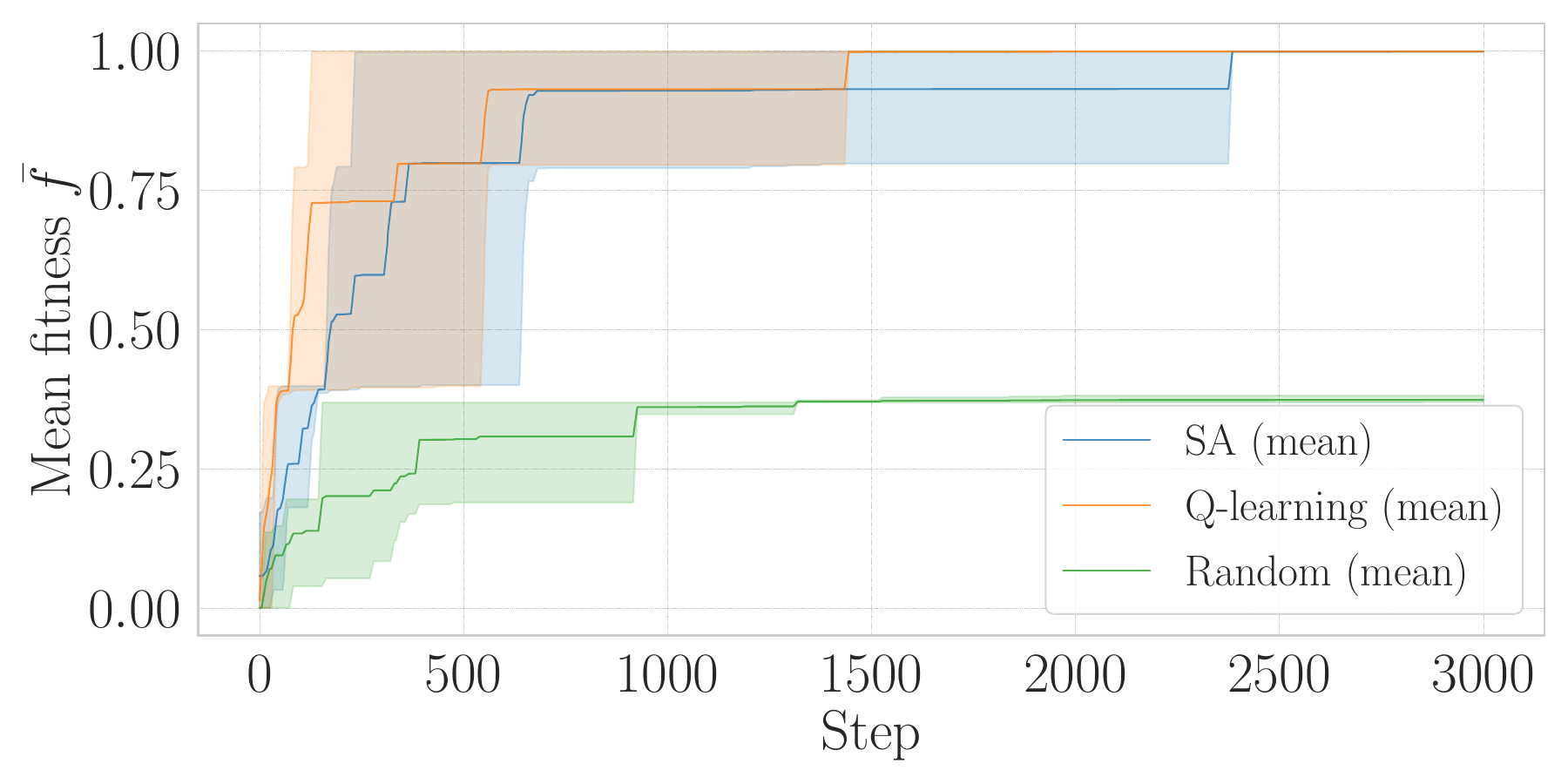}}
\caption{Average fitness $\bar f$ for different learning algorithms, aggregated over $N=10$ independent runs. Each curve shows the mean across runs, the shaded band indicates the corresponding min-max.}
	\label{fig:sim_experience}
\end{figure}

Both Q-learning and SA rapidly acquire the full skill set, achieving near-optimal performance, whereas the random policy fails to develop meaningful behavior (Fig.~\ref{fig:sim_experience}).

Q-learning converges fastest, followed closely by SA. This indicates that the intrinsic motivation paired with the curriculum mechanism based on the \gls{kg} in \gls{dsm} can effectively guide both learners toward the relevant search spaces. The random baseline stagnates far below the threshold for meaningful skill acquisition. This is because more complex skills, such as higher-order movement skills, are the primary bottlenecks. Fitness in these skills is based on a specific target (displacement or angle), and uniform action sampling almost never achieves this target, while SA and Q-learning both master these skills.

\subsubsection{Intrinsic Motivation Dynamics}

\begin{figure}[htbp]
	\centerline{\includegraphics[width=0.5\textwidth]{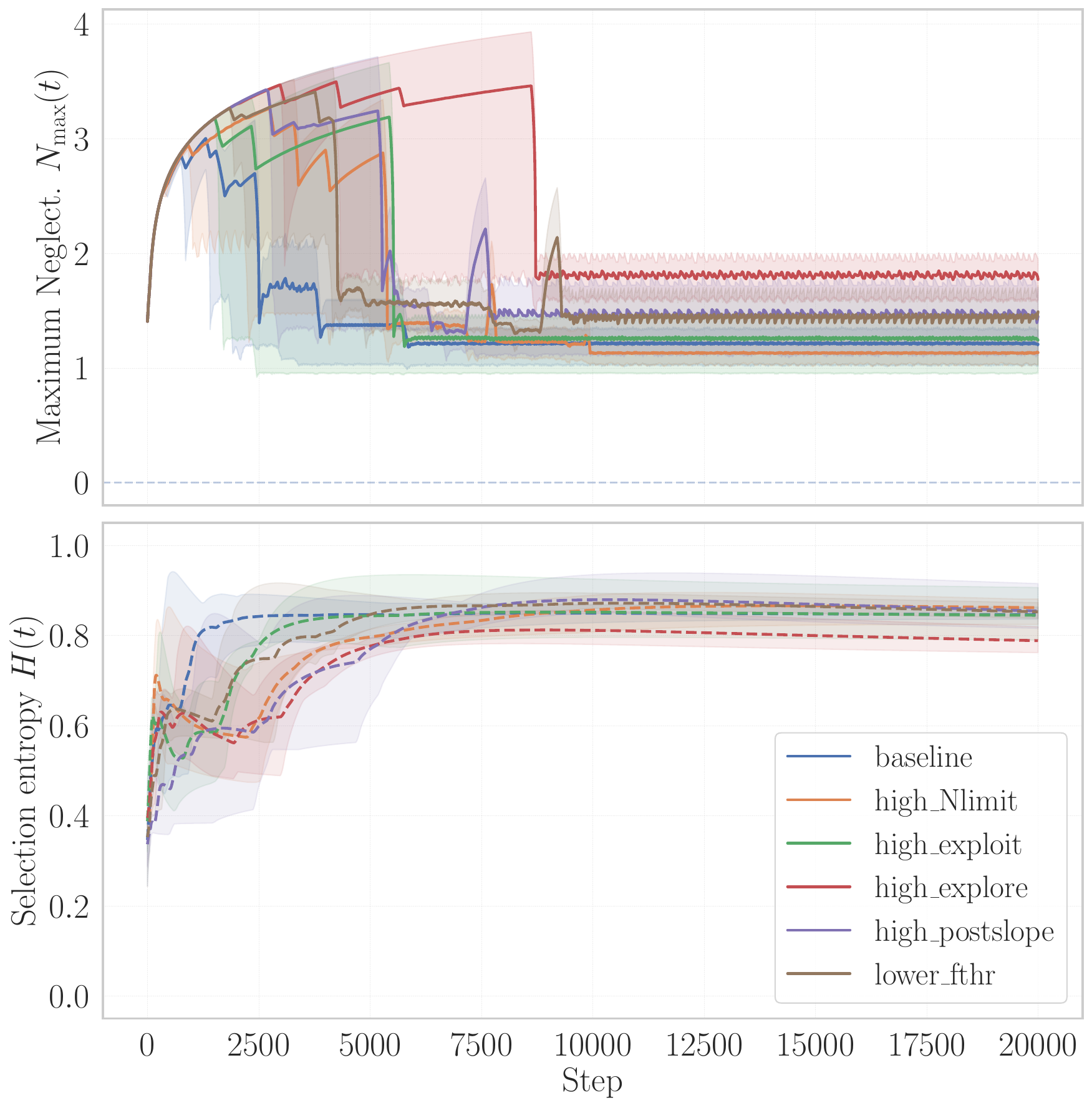}}
\caption{Intrinsic Motivation Dynamics across $N=10$ independent runs. Top: maximum neglect $N_{\max}(t)$; bottom: selection entropy $H(t)$. Curves show the mean across runs, with shaded bands illustrating the across-run variability.}
	\label{fig:im_health}
\end{figure}

Fig.~\ref{fig:im_health} shows the corresponding IM analysis. High exploration (\emph{high\_explore}, large $\gamma$) leads to high selection entropy but also extreme maximum neglect, indicating that some skills are ignored for long periods. High exploitation (\emph{high\_exploit}, large $\beta$) maintains low neglect but slightly reduces coverage. The baseline configuration maintains both high entropy ($\approx 0.83$) and low maximum neglect ($\approx 1.37$), indicating a balanced and stable development process.

Table~\ref{tab:im_scores} shows the IM Score for each configuration. Since the score combines the learning quality ($\bar f$) with scheduling stability ($\overline{N}_{\max}$ and $\bar H$), it directly mirrors the trade-off between fast skill acquisition and robust development behavior.

The baseline configuration achieves the highest IM rating, which is also evident in Fig.~\ref{fig:im_health}. It incorporates fast, high-quality learning with stable and well-balanced skill scheduling.

The \emph{high\_exploit} configuration reduces maximum neglect but also slightly lowers selection entropy, indicating an early focus on a narrow development schedule. The \emph{high\_Nlimit} maintains exploration diversity but allows for higher neglect by focusing too much on novelty-oriented exploration.

The \emph{lower\_fthr} setting leads to early mastery but increases neglect, where some skills are effectively never selected again on the timescale of learning. On the other hand, \emph{high\_postslope} keeps mastered skills highly attractive, causing the scheduler to repeatedly revisit a few skills and thereby reducing the stability of the developmental schedule.

Finally, the \emph{high\_explore} performs worst, as it focuses on extreme novelty, which causes the IM scheduler to neglect previously learned skills, resulting in long-term starvation of many skills despite high overall entropy.

Table~\ref{tab:im_scores} summarises the resulting scores. For completeness, we present in Appendix~\ref{appendix:im-pareto} a Pareto diagram of the trade-off between mean fitness, maximum neglect, and entropy across IM configurations, which corroborates that the baseline configuration offers the most balanced compromise.
\begin{table}[h]
\centering
\caption{IM configuration comparison. A higher IM Score is better.}
\label{tab:im_scores}
\renewcommand{\arraystretch}{1.4}
\begin{tabular}{lcccc}
\hline
\textbf{IM}
& $\bar f$
& $\bar H$
& $\overline{N}_{\max}$
& \textbf{IM Score} \\
\hline
baseline        & 0.976 & 0.834 & 1.38 & $\mathbf{0.673 \pm 0.060}$ \\
high\_exploit   & 0.962 & 0.817 & 1.50 & $0.628 \pm 0.089$ \\
high\_Nlimit    & 0.943 & 0.808 & 1.53 & $0.600 \pm 0.086$ \\
lower\_fthr     & 0.922 & 0.827 & 1.75 & $0.525 \pm 0.054$ \\
high\_postslope & 0.903 & 0.803 & 1.75 & $0.506 \pm 0.075$ \\
high\_explore   & 0.932 & 0.763 & 2.27 & $0.403 \pm 0.048$ \\
\hline
\end{tabular}
\vspace{1mm}
\caption*{\footnotesize Means over $n=10$ runs; IM Score reported as mean $\pm$ standard deviation across runs.}
\end{table}

These results demonstrate that \gls{dsm} is highly sensitive to intrinsic motivation parameterization. Increasing the novelty destabilizes development through excessive neglect, whereas stronger exploitation accelerates early learning at the cost of reduced diversity. The baseline configuration offers the most robust development schedule. This indicates that IM scheduling plays a stability-critical role in shaping developmental dynamics.

\section{Conclusion}

In this work, we introduced \gls{dsm}, an advanced developmental skill method specifically designed for resource-constrained millirobots, bridging the domains of tiny robot learning and cognitive developmental robotics. \gls{dsm} enables autonomous, self-directed skill development without an externally specified task sequence while operating within a predefined knowledge graph. It combines intrinsic motivation, fitness-based evaluation, and structured knowledge representation through a hierarchical knowledge graph and kinematic reasoning.

We demonstrated that a millirobot with a volume of only \qty{36}{\centi\meter\cubed}, running on a Raspberry Pi Pico 32-bit microcontroller (RP2040), can progress from atomic motion patterns to complex geometric behaviors within 15 minutes, despite operating within just \qty{9}{\kilo\byte} of memory. This is achieved through curriculum-based learning and an intrinsic motivation model balancing novelty, progress, and difficulty.

To assess generality and robustness, we conducted a complementary simulation-based analysis. This allowed systematic evaluation across different learning algorithms and intrinsic motivation configurations. The results highlight that the developmental dynamics and stability of skill acquisition are highly sensitive to motivation scheduling and parameterization.

We showed that the agent efficiently developed core motion skills, adapted to physical changes like added weight, and maintained performance through continuous fitness-based evaluation. The system shows robustness in maintaining previously acquired skills while pursuing new competencies, supporting the concept of lifelong learning. These results demonstrate the potential of flexible, self-directed robotic agents that can learn throughout their lifetime in dynamic environments with limited resources. 

Limitations of the current approach include that fitness-based monitoring can relearn skills after moderate changes (e.g., added weight), but compensation for sensor and actuator error only applies when deviations remain visible in the sensor stream. Currently, the experiments are set up in a closed-loop using an external camera instead of onboard localization, which reduces system autonomy and limits transferability to real-world deployments where the camera is unavailable or unreliable. In addition, the hierarchical \gls{kg} consists of manually specified generic prior knowledge and is kept fixed at runtime in this study, limiting skill availability and curriculum progression to predefined structures rather than demonstrating scalable graph extension.

Future work will explore richer sensing, higher-level planning, lightweight on-device neural network learning, evaluation in more complex scenarios, and over-the-air structural updates to enable additional skill discovery. 

\appendices
\section{Intrinsic Motivation}
\label{appendix:intrinsic-motivation}
For compactness, we abbreviate the novelty, progress, and difficulty functions as $n(o, t)$, $p(o, t)$, and $d(o, t)$, respectively. The \gls{im} $m(o,t)$ of an agent to pursue a masterable skill with $o \in M_{O}$ as the product of three factors with:

\begin{equation*}
	\begin{aligned}
		&n(o, t) = \\
        &
        \begin{cases}
         N_{\text{init}} & t = t_0 \\
        n(o, t_{-1}) \cdot (1 - \beta) & \text{if executed at} \; t \\
        n(o, t_{-1}) + \gamma (N_{\text{limit}} - n(o, t_{-1})) & \text{otherwise}
        \end{cases}   \\
	\end{aligned}
\end{equation*}

\noindent where $\beta$ is the decay rate, $\gamma$ is the growth rate, $N_\text{init}$ is the initial and $N_\text{limit}$ is the maximum novelty value.

\begin{equation*}
	\begin{aligned}
    &p(o, t) = \\
	&
    \begin{cases}
        f(o, t) \cdot \frac{p_{\text{scale}}}{f_{\text{threshold}}} + p_{\text{offset}} & \text{if } f(o, t) < f_{\text{threshold}} \\
        1 - f(o, t) \cdot p_{\text{offset}} & \text{if } f(o, t) \geq f_{\text{threshold}}
    \end{cases}
	\end{aligned}
\end{equation*}
where $p_{\text{scale}}$ and $p_{\text{offset}}$ are scaling and offset parameters for progress.

\begin{equation*}
	\begin{aligned}
    d(o, t) = \log_2\left(2 + \sum_{o_k \in \mathrm{pred}(o)} f(o_k, t) \right),
	\end{aligned}
\end{equation*}
where $f(o_k, t)$ is the fitness of each prerequisite skill $o_k \in \mathrm{pred}(o)$ at time $t$ and the set of prerequisites follows with:

\begin{equation*}
\begin{aligned}
    \mathrm{pred}(o) = \{ o_i \in O \mid\
      &o_i \rightarrow o \in U\ \lor\\
      &\exists o_j \in \mathrm{pred}(o):\ o_i \rightarrow o_j \in U \}.
\end{aligned}
\end{equation*}

Finally, at every discrete time step $t$, the agent selects the next skill to explore by maximizing the motivation function:
\begin{equation}
	\begin{aligned}
    o_{\text{next}}(M_o, t) := \underset{o \in M_o}{\arg\max}\, m(o, t),
	\end{aligned}
\end{equation}
where $M_o$ is the available skill pool.

\section{Experience}
\label{appendix:experience}
We define the experience $E(o, t)$ to indicate whether a skill $o \in M_o$ has been learned at time $t$.
\begin{equation*}
    E(o, t) =
        \begin{cases}
            1, & \text{if } f(o, t) \geq f_{\text{threshold}}, \\
            0, & \text{otherwise}
        \end{cases}
\end{equation*}

The total accumulated experience of all skills at time $t$ is defined as:
\begin{equation}
    E_t = \sum_{o_k \in M_o} E(o_k, t)
\end{equation}

This metric is used to reflect the overall progress of the agent.

\section{Intrinsic Motivation Health Metrics}
\label{appendix:im-metrics}

\paragraph{Maximum Neglect.}
Let $R_k(t)$ be the number of steps since skill $k$ was last selected. The worst-case neglect is
\begin{equation}
N_{\max}(t)=\log_{10}\!\big(\max_k R_k(t)+\varepsilon\big),\qquad \varepsilon=10^{-3}.
\end{equation}

\paragraph{Selection Entropy.}
With cumulative counts $C_k(t)$, the selection probability is
\begin{equation}
p_k(t)=\frac{C_k(t)}{\sum_{j=1}^{K}C_j(t)}.
\end{equation}
The normalized Shannon entropy is
\begin{equation}
H(t)=-\frac{1}{\log K}\sum_{k:p_k(t)>0} p_k(t)\log p_k(t).
\end{equation}

Here $H(t)\!\in\![0,1]$ measures coverage of skill sampling, while $N_{\max}(t)$ captures long-term skill neglect.

\section{Intrinsic Motivation Pareto Analysis}
\label{appendix:im-pareto}

\begin{figure}[hb]
    \centering
    \includegraphics[width=0.45\textwidth]{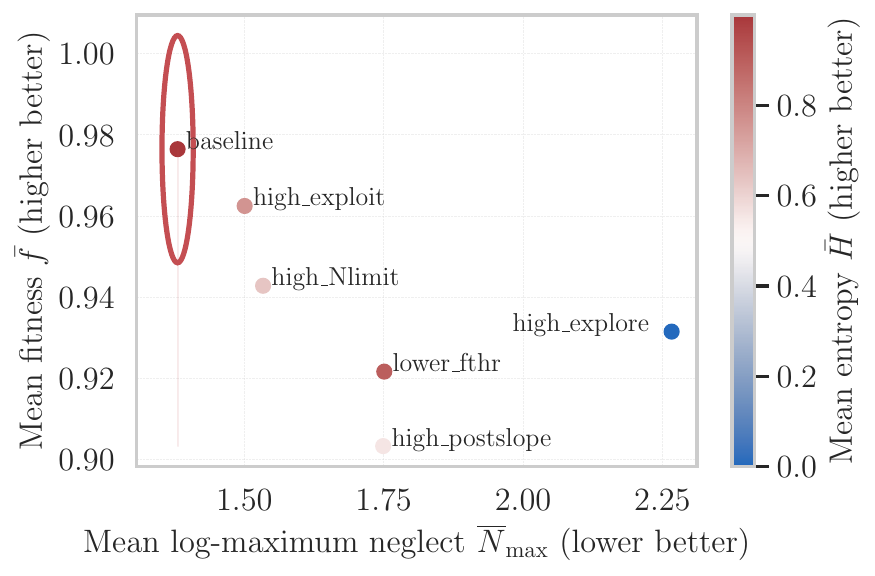}
    \caption{Pareto diagram with trade-off between mean fitness $\bar f$ (higher is better), mean log-maximum neglect $\overline{N}_{\max}$ (lower is better), and mean entropy $\bar H$ (higher is better) for the IM configurations from Table~\ref{tab:im_scores}. The 2D projection shows $\bar f$ versus $\overline{N}_{\max}$, with $\bar H$ encoded by color and Pareto-efficient configurations highlighted by a red outline.}
    \label{fig:im_pareto}
\end{figure}

We analyze the trade-off between learning performance, long-term neglect, and entropy across IM configurations. Figure~\ref{fig:im_pareto} shows a Pareto-style view with one point per configuration, using the configuration-level means of $\bar f$, $\overline{N}_{\max}$, and $\bar H$ reported in Table~\ref{tab:im_scores}.

\bibliographystyle{IEEEtran}
\bibliography{_bib}

\end{document}